\documentclass[sigconf]{acmart}
\AtBeginDocument{%
  }

\setcopyright{none}
\copyrightyear{2026}
\acmYear{2026}
\acmDOI{}

\acmConference[Preprint]{Preprint}{2026}{}
\acmISBN{}

\renewcommand\footnotetextcopyrightpermission[1]{}
\usepackage{stfloats}
\usepackage{listings}
\usepackage{graphicx}
\usepackage{tcolorbox}
\tcbuselibrary{breakable}

\lstdefinelanguage{json}{%
  basicstyle=\ttfamily\footnotesize,
  showstringspaces=false,
  breaklines=true,
  literate=
   *{0}{{{0}}}{1}{1}{{{1}}}{1}{2}{{{2}}}{1}{3}{{{3}}}{1}{4}{{{4}}}{1}
    {5}{{{5}}}{1}{6}{{{6}}}{1}{7}{{{7}}}{1}{8}{{{8}}}{1}{9}{{{9}}}{1}
    {:}{{{:}}}{1}{,}{{{,}}}{1}{\{}{{{\{}}}{1}{\}}{{{\}}}}{1}
    {[}{{{[}}}{1}{]}{{{]}}}{1},
}

\usepackage{dsfont}   
\usepackage{algorithm}
\usepackage{algpseudocode}  
\usepackage{amsmath}
\usepackage{multicol}
\usepackage{enumitem}
\newcommand{\Crop}{\textsc{Crop}}
\newcommand{\VLMGenerate}{\textsc{VLM\_Inference}}
\newcommand{\ParseSeq}{\textsc{ParseSequence}}
\newcommand{\LoadTemplate}{\textsc{LoadTemplate}}
\newcommand{\VectorMatching}{\textsc{DirectionalVectorMatching}}
\newcommand{\Align}{\textsc{GeometricAlign}}
\newcommand{\RDKitMol}{\textsc{RDKit.MolFromGraph}}
\newcommand{\RDKitSMILES}{\textsc{RDKit.MolToSmiles}}

\usepackage{times}
\usepackage{latexsym}

\usepackage[T1]{fontenc}

\usepackage[utf8]{inputenc}
\usepackage{microtype}

\usepackage{graphicx}
\usepackage{array}
\usepackage{multirow}
\usepackage{tcolorbox}
\usepackage{amsmath} 
\usepackage{booktabs}
\usepackage{mathtools}
\usepackage{amsfonts}
\usepackage{seqsplit}

\newcommand{\ie}{\emph{i.e.}}
\newcommand{\eg}{\emph{e.g.}}
\newcommand{\OURS}{{ChemVA}}

\begin{document}

\title{\OURS{}: Advancing Large Language Models on Chemical Reaction Diagrams Understanding}

\author{Mingyang Rao}
\authornote{Both authors contributed equally to this research.}
\affiliation{%
  \institution{College of Computer Science and Technology, Zhejiang University}
  \city{Hangzhou}
  \state{Zhejiang}
  \country{China}}
\email{myR001@zju.edu.cn}

\author{Kehua Feng}
\authornotemark[1]
\affiliation{%
  \institution{College of Computer Science and Technology, Zhejiang University}
  \city{Hangzhou}
  \state{Zhejiang}
  \country{China}}
\email{kehuafeng@zju.edu.cn}

\author{Zhihui Zhu}
\affiliation{%
  \institution{ZJU-Hangzhou Global Scientific and Technological Innovation Center, Zhejiang University}
  \city{Hangzhou}
  \state{Zhejiang}
  \country{China}}
\email{Zhihui.Zhu01@outlook.com}

\author{Jiangzhen Fu}
\affiliation{%
  \institution{Department of Chemistry, Fudan University}
  \state{Shanghai}
  \country{China}}
\email{23110220019@m.fudan.edu.cn}

\author{Hao Yu}
\affiliation{%
  \institution{Department of Chemistry, Fudan University}
  \state{Shanghai}
  \country{China}}
\email{hao_yu@fudan.edu.cn}

\author{Keyan Ding}
\authornote{Corresponding author.}
\affiliation{%
  \institution{ZJU-Hangzhou Global Scientific and Technological Innovation Center, Zhejiang University}
  \city{Hangzhou}
  \state{Zhejiang}
  \country{China}}
\email{dingkeyan@zju.edu.cn}

\author{Huajun Chen$^\dag$}
\affiliation{%
  \institution{College of Computer Science and Technology, Zhejiang University}
  \city{Hangzhou}
  \state{Zhejiang}
  \country{China}}
\email{huajunsir@zju.edu.cn}

\renewcommand{\shortauthors}{Rao et al.}

\begin{abstract}
While Large Language Models (LLMs) have revolutionized scientific text processing, they exhibit a significant capability gap when interpreting chemical reaction diagrams. We identify two fundamental bottlenecks restricting current systems: a \textit{Visual Deficit}, where generic vision encoders struggle to resolve the strict topological connectivity of dense molecular graphs, and a \textit{Semantic Disconnect}, where standard linear strings (\eg, SMILES) fail to effectively activate the model's latent chemical reasoning. To bridge these gaps, we propose the \textbf{Chemical Visual Activation (\OURS{})} framework, which employs a novel \textit{Visual Anchor} mechanism to ground functional groups via hybrid-granularity detection, followed by a semantic alignment approach that translates visual features into entity names to maximize knowledge activation in LLMs. 
We evaluate our approach on \textbf{OCRD-Bench}, a newly constructed dataset featuring dense visual-semantic contexts and comprehensive reaction coverage to evaluate the full spectrum from recognition to reasoning. Extensive experiments on OCRD-Bench demonstrate that \OURS{} achieves 92.0\% structural recognition accuracy. By bridging visual and semantic bottlenecks, our framework delivers a consistent performance gain of approximately 20 percentage points across 9 diverse LLMs, enabling open-weights models to rival proprietary SOTA systems in complex chemical reasoning tasks. 


\end{abstract}



\maketitle


\begin{figure}[t]
    \centering    \includegraphics[width=1.0\linewidth]{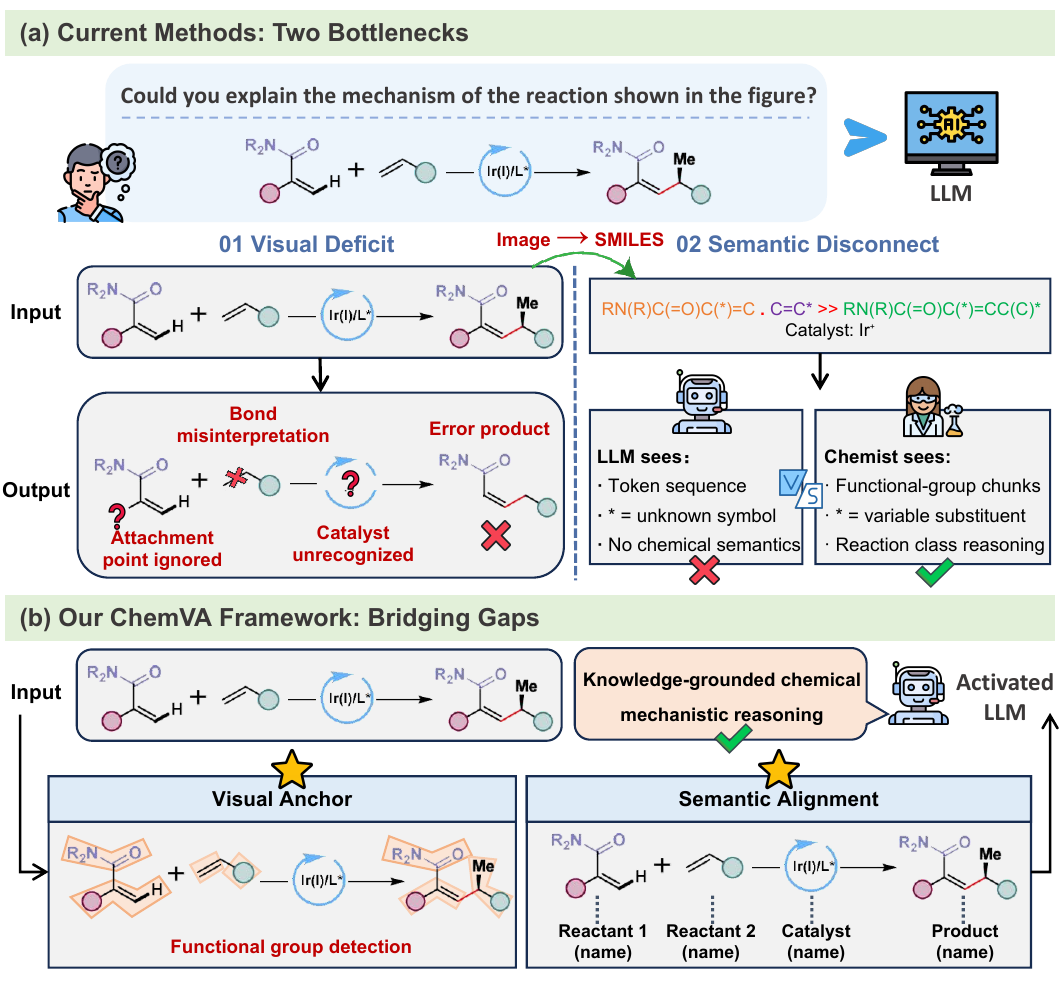}
    \caption{Overview of limitations in LLM-based chemical reaction diagram understanding and the proposed \OURS{} framework. 
    (a) Existing pipelines exhibit a Visual Deficit in Reaction Diagram Parsing and a Semantic Disconnect under SMILES-based representations. (b) \OURS{} bridges these gaps via functional-group--grounded Reaction Diagram Parsing and Semantic Activation for downstream reasoning.}
    \label{fig:intro_gap}
    \vspace{-1em}
\end{figure}

\section{Introduction}\label{sec1}

The integration of Large Language Models (LLMs) into scientific discovery pipelines has accelerated progress in ``AI for Science'', especially in domains that require complex reasoning, such as organic chemistry \cite{boiko2023autonomous, zhang2024igniting}. General-purpose models such as GPT-5 and Gemini2.5-pro have shown strong performance on text-centric chemical tasks, including synthesis planning and property prediction \cite{Bran2025Chemical,Gu2025MolRAG,Pharmacy2025Gemini}. However, the primary modality of chemical knowledge in the literature remains visual, \ie, molecular structures and reaction diagrams, and current state-of-the-art models still exhibit a substantial gap on this modality \cite{li2025chemvlm, guo2023what}. This gap can be traced to two fundamental bottlenecks: a \textit{Visual Deficit} in structural perception and a \textit{Semantic Disconnect} in knowledge activation (Figure~\ref{fig:intro_gap}).

First, despite rapid progress in vision-language models, a critical \textbf{Visual Deficit} remains in the processing of chemical images. Chemical structures are dense graph objects whose meaning is determined by topological connectivity and stereochemical geometry, rather than by the global semantic cues that CLIP-style encoders typically emphasize \cite{morin2023molgrapher,staker2019molecular}. Prior evaluations \cite{qian2023molscribe, clevert2021img2mol} show that general-purpose models often hallucinate atoms, confuse bond types, and degrade under rotations or visually crowded graphs (\ie, molecules with dense functional groups or complex ring systems). In contrast to natural images, where mild distortions may be tolerable, even a single local mistake in a chemical diagram (\eg, interpreting a wedge bond as a dashed bond) can change molecular identity and invalidate downstream analysis. Existing Optical Chemical Structure Recognition (OCSR) tools, including MolScribe\cite{qian2023molscribe}, Decimer\cite{rajan2020decimer,decimer2023}, and related methods\cite{chen2025chemminer,oldenhof2021self,shah2025multimodal,wang2025gtr,zhuang2025doc2sar}, have improved image-to-graph translation, but the dominant formulation remains \textit{atom-level} prediction that represents molecules as sequences of nodes and edges. This formulation does not capture \textit{substructure-level} perception, where functional groups function as stable semantic units that guide expert interpretation.

Second, even when structure recognition yields a correct linear representation such as SMILES, downstream reasoning in LLMs can still be hindered by what we term a \textbf{Semantic Disconnect}. This term refers to the mismatch between chemically valid symbolic encodings and the semantic cues that LLMs rely on during inference. While SMILES provides a precise specification of molecular graphs, it is largely a non-semantic character sequence for a language model and does not reliably evoke the functional concepts that support chemical reasoning. Prior evidence indicates that SMILES alone often fails to effectively ``activate'' latent chemical knowledge in LLMs \cite{edwards2022translation}. Conversely, chemical entity names (\eg, IUPAC names\cite{heller2015inchi}) provide explicit descriptors that more consistently elicit knowledge about reactivity and synthesis \cite{runcie2025reasoning, diao2023macfrag}. Therefore, pipelines that follow ``Image $\to$ SMILES $\to$ Reasoning'' can underutilize the reasoning capacity of LLMs, because the intermediate representation is not aligned with the semantic abstractions that guide inference.

To bridge these gaps, we propose Chemical Visual Activation (\textbf{\OURS{}}), a two-phase framework that comprises \textbf{Reaction Diagram Parsing} and \textbf{Semantic Activation}. Motivated by the ``chunking'' mechanism in human cognition \cite{gobet2001chunking}, the Reaction Diagram Parsing phase uses functional-group-grounded vision-language model (FG-VLM) to ground functional groups as \textit{Visual Anchors} through reaction deconstruction and hybrid-granularity visual anchor modeling, enabling topology-faithful structural reconstruction. The Semantic Activation phase then performs semantic alignment by resolving the recognized structures into entity names, which are used to construct semantically grounded prompts that more effectively elicit chemical knowledge in downstream LLMs.
To overcome the limitations of existing datasets, which typically emphasize either structural recognition or text-only reasoning, we construct \textbf{OCRD-Bench}, a unified multimodal benchmark derived from graduate-level chemistry entrance exams.
Experiments on OCRD-Bench show that \OURS{} consistently outperforms strong baselines. In particular, FG-VLM reduces error accumulation that is common in specialized OCSR pipelines and mitigates structural hallucinations in general-purpose multimodal models (\eg, spurious atoms/bonds and incorrect connectivity). In addition, \textbf{Semantic Activation} addresses the Semantic Disconnect and improves downstream reasoning by providing entity-name--grounded prompts that better align with the semantic abstractions used by LLMs.

Our contributions can be summarized as follows:
\begin{itemize}
\item We propose \OURS{}, a two-phase framework that bridges the \textit{Visual Deficit} and \textit{Semantic Disconnect} through \textbf{Reaction Diagram Parsing} and \textbf{Semantic Activation}, improving diagram understanding and downstream reasoning in chemical MLLMs.
\item We develop FG-VLM, the first hybrid-granularity OCSR model that grounds functional groups as \textit{Visual Anchors} and reconstructs topology-faithful molecular graphs via Directional Vector Matching, going beyond purely atom-level serialization.
\item We construct OCRD-Bench, a challenging multimodal benchmark derived from graduate-level entrance exams, with broad reaction coverage and a hierarchical evaluation protocol that spans visual perception, chemical knowledge, and mechanistic reasoning.
\item We show that \OURS{} consistently improves performance across a wide range of LLMs, yielding substantial gains on complex reasoning tasks.
\end{itemize}

\section{Related Work}
\subsection{Methods for Chemical Reaction Diagram Understanding}

Prior research on chemical reaction diagram understanding largely falls into two lines: (i) symbolic serialization methods that ``digitize'' structures from images, and (ii) multimodal LLMs that attempt to reason directly from visual inputs.

A long-standing direction is {symbolic serialization} via OCSR, which converts raster diagrams into machine-readable symbolic forms such as molecular graphs or SMILES. Representative systems (\eg, MolScribe \cite{qian2023molscribe}, RxnScribe \cite{rxnscribe2024}, OpenChemIE\cite{fan2024openchemie}, and DECIMER \cite{rajan2020decimer,decimer2023}, ReactionDataExtractor1.0/2.0 \cite{wilary2021reactiondataextractor,wilary2023reactiondataextractor}) typically operate at {atom-level} granularity and formulate recognition as structured prediction (graph prediction or graph-to-text encodings \cite{yu2024g2t}). While these models can achieve strong structural accuracy on clean inputs, they primarily function as semantics-agnostic digitizers: the output symbols are chemically valid but do not explicitly expose higher-level semantic units (\eg, functional groups) that are critical for robust interpretation under rotation/crowding and for downstream reasoning. 

Recent work couples vision encoders with LLMs to enable end-task reasoning. \emph{Implicit alignment} approaches (\eg, ChemVLM \cite{li2024chemvlm}) align generic vision backbones with chemical text through instruction tuning, but often suffer a {visual deficit} when faced with dense molecular graphs, leading to structural hallucinations relative to specialized OCSR tools. Another line, \emph{external retrieval} (\eg, MolMole \cite{research2025molmole}, Doc2SAR \cite{zhuang2025doc2sar}), grounds LLM outputs by extracting or retrieving molecular facts from documents and databases; this can improve factuality for lookup-like tasks, but it relies on external resources rather than strengthening intrinsic visual understanding and generative reasoning. 

Our framework bridges these paradigms with (i) a hybrid-granularity visual module that grounds \emph{functional groups} as {visual anchors} for more robust structure understanding beyond pure atom-level serialization, and (ii) a semantic alignment workflow that translates grounded visual evidence into \emph{entity names} to maximize knowledge activation in downstream LLMs, without requiring database retrieval.

\subsection{Benchmarks for Chemical Reaction Diagram Understanding}
Existing benchmarks typically isolate specific stages of the chemical reaction diagram understanding pipeline, failing to evaluate end-to-end capabilities.
{Recognition-centric benchmarks} (\eg, standard USPTO subsets\cite{oldenhof2021self}) focus exclusively on Image-to-SMILES translation accuracy, ignoring downstream chemical validity.
{Text-centric reasoning benchmarks}, such as {SciBench} \cite{wang2024scibench} and {ChemBench} \cite{zhang2024chemllm}, assess knowledge but primarily use text inputs, bypassing visual perception challenges.
While recent multimodal benchmarks like {MMCR} \cite{li2024chemvlm} and {MaCBench} \cite{alampara2024macbench} introduce visual reasoning, they are predominantly limited to {multiple-choice questions} or document-level element detection. Crucially, despite recent progress in molecular and reaction comprehension tasks\cite{runcie2025chemiq,reactbench2025,xu2025omebench,moleculariq2026}, there remains a significant scarcity of benchmarks that specifically bridge the gap between visual reaction recognition and deep mechanistic reasoning \cite{yue2024mmmu,lu2022learn}.

Derived from graduate-level entrance examinations, our OCRD-Bench is the first benchmark to demand end-to-end generative reasoning, requiring models to independently parse complex reaction diagrams and generate multi-step mechanism explanations.

\section{Method}
\label{sec:method}

\begin{figure*}[t]
    \centering
    \includegraphics[width=1.0\linewidth]{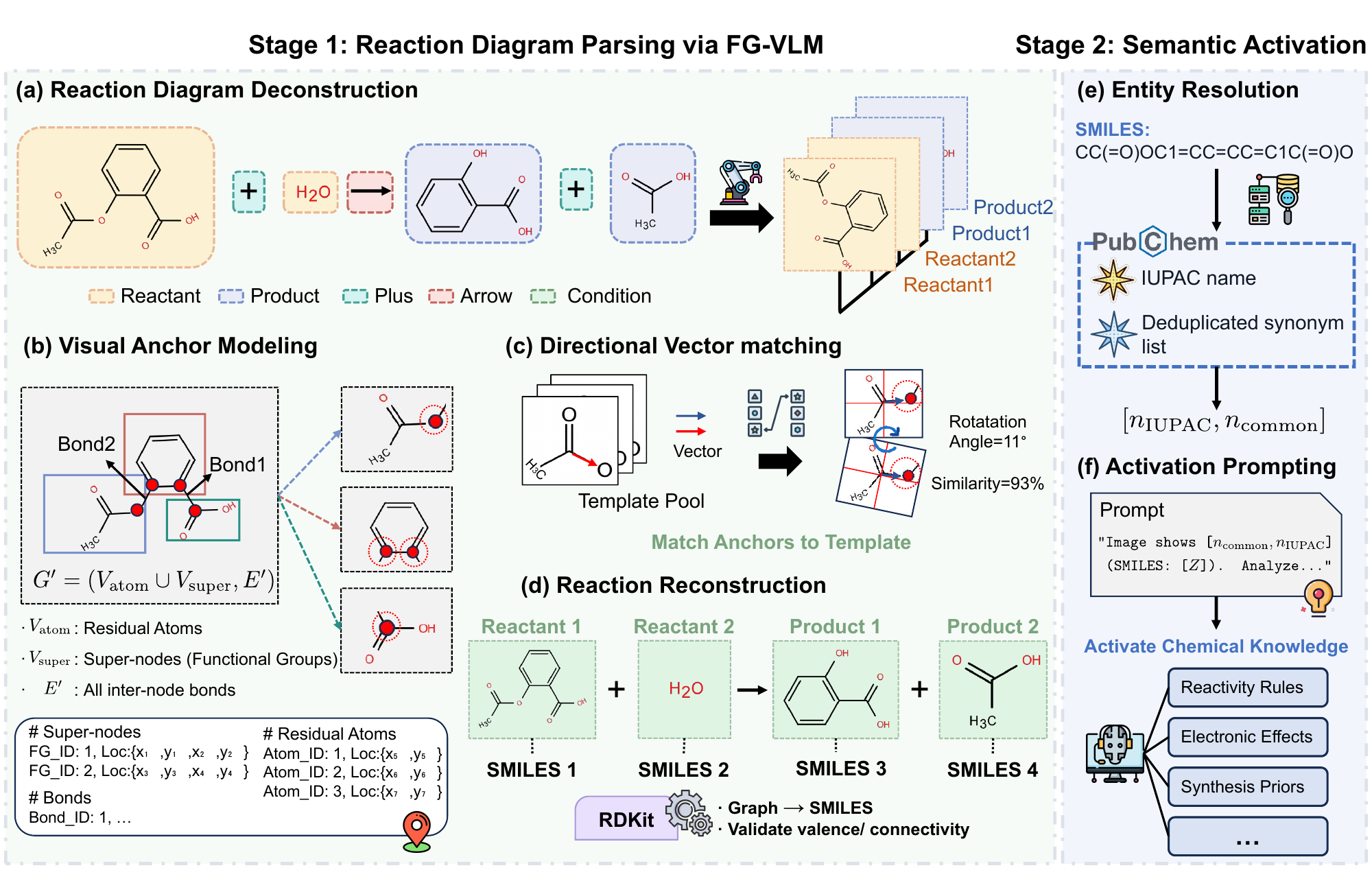}
    \caption{Overview of the ChemVA framework.
    \textbf{Stage 1: Reaction Diagram Parsing.} (a) \textbf{Reaction Diagram Deconstruction}: FG-VLM localizes diagram elements (\eg, reactants, conditions, arrows) and crops molecular regions. (b) \textbf{Visual Anchor Modeling}: FG-VLM predicts a hybrid-granularity graph (functional groups and residual atoms) with bonds, and regresses group attachment anchors as pixel keypoints. (c) \textbf{Directional Vector Matching}: Group templates are oriented and attached to anchors to instantiate a topology-faithful molecular graph. (d) \textbf{Reaction Reconstruction}: The recovered graphs are converted to SMILES and parsed into a structured reaction sequence.
    \textbf{Stage 2: Semantic Activation.} (e) \textbf{Entity Resolution}: SMILES are resolved to IUPAC/common names via PubChem. (f) \textbf{Activation Prompting}: The names are used to construct semantically grounded (name-first) prompts that more reliably elicit chemical knowledge for downstream reasoning.}
    \label{fig:architecture}
\end{figure*}

This section presents the proposed \textbf{\OURS{}} framework for improving LLM understanding of chemical reaction diagrams. As shown in Figure~\ref{fig:architecture}, given a reaction image, \OURS{} reconstructs a topology-faithful molecular graph and its reaction semantics through two main phases:
(1) \textbf{Diagram-to-Reaction Parsing} (Sec.~\ref{ssec:fgvlm_pipeline}), where we propose \textbf{FG-VLM} to progressively recover the structured representation by performing (i) \textit{reaction deconstruction}, (ii) \textit{hybrid-granularity visual anchor modeling} for node and anchor prediction, and (iii) \textit{structural reconstruction} via directional vector matching, followed by (iv) \textit{reaction reconstruction} to assemble the complete reaction sequence; and
(2) \textbf{Semantic Activation} (Sec.~\ref{ssec:activation}), which translates the recognized structure into semantically grounded prompts to elicit chemical knowledge from LLMs for downstream reasoning.
Finally, we detail the \textbf{data construction and training} procedure (Sec.~\ref{sec:data_and_training}) used to equip FG-VLM with hybrid-granularity perception via a curriculum-style strategy.

\subsection{Reaction Diagram Parsing via FG-VLM}
\label{ssec:fgvlm_pipeline}

Given a chemical reaction diagram image $I$, our goal is to recover a structured reaction representation that supports downstream chemical visual reasoning. Prior approaches typically cast diagram understanding as atom-level sequence generation, \ie, emitting an ordered list of atoms $V_{\text{atom}}$ and their bonds \cite{qian2023molscribe}. However, organic molecules contain recurring rigid motifs whose semantics are determined at the substructure level; purely atom-level generation therefore produces long outputs and is prone to local errors that propagate to global topology.

To address this, we present functional-group-grounded vision-language model (FG-VLM), which performs end-to-end diagram-to-reaction parsing. Specifically, FG-VLM is designed to (i) detect and recognize rigid chemical substructures (functional groups) and residual atoms in $I$, (ii) reconstruct their connectivity to form a topology-faithful molecular graph, and (iii) recover the full reaction structure to derive canonical text representations (\eg, SMILES). The detailed prompt specifications and instruction templates for each stage are provided in Appendix~\ref{sec:appendix_prompts}.

\subsubsection{Reaction Diagram Deconstruction.}
\label{ssec:deconstruction}
The first bottleneck in processing reaction diagrams is disentangling crowded visual elements. As shown in the top panel of Figure~\ref{fig:architecture}, we employ FG-VLM as a layout parser to perform \textit{Reaction Deconstruction}.
Given the raw image $I$, the model detects bounding boxes for all semantic constituents: $\mathcal{O} = \{ (c_i, \mathbf{b}_i) \}_{i=1}^N$, where $N$ is the number of objects and classes $c_i \in \{\text{Reactant},\allowbreak \text{Product},\allowbreak
  \text{Condition},\allowbreak \text{Arrow},\allowbreak \text{Plus}\}
$. Unlike generic object detectors, our model is trained to implicitly understand chemical syntax.
Based on the detected boxes $\mathbf{b}_i$, we perform a \textsc{Crop} operation to generate a set of isolated molecular sub-images, denoted as $\mathcal{I}_{crops} = \{ I_{crop}^{(i)} \mid c_i \in \{\text{Reactant},\text{Product}\} \}$.
These isolated views serve as the input for the subsequent fine-grained structure recognition.

\subsubsection{Hybrid-Granularity Visual Anchor Modeling.}
\label{ssec:visual_anchor}
For each cropped molecule image $I_{crop} \in \mathcal{I}_{crops}$, we aim to reduce atomic redundancy while preserving exact topology. We represent the molecule as a \emph{hybrid-granularity graph} $G'=(V_{\text{atom}} \cup V_{\text{super}}, E')$, where $V_\text{super}$ denotes predefined semantic super-nodes (\eg, Phenyl, Carboxyl), including terminal groups as well as bridging/scaffold groups with multiple attachment sites, and $V_\text{atom}$ contains the remaining residual atoms. The edge set $E'$ encodes bonds among all nodes in $V_{\text{atom}} \cup V_{\text{super}}$, including Atom--Atom, Atom--SuperNode, and SuperNode--SuperNode bonds.

\paragraph{Hybrid-Granularity Nodes and Bonds.}
Given $I_{crop}$, FG-VLM predicts a unified node set $V = V_{\text{atom}} \cup V_{\text{super}}$ and a bond set $E$.
Each super-node $g\in V_{\text{super}}$ is represented as $(id_g, name_g, \mathbf{b}_g)$, where $id_g$ is a unique node ID, $name_g$ specifies the functional-group type, and $\mathbf{b}_g=(x_{g,1},y_{g,1},x_{g,2},y_{g,2})$ is its bounding box.
Each residual atom $a\in V_{\text{atom}}$ is represented as $(id_a, elem_a, \mathbf{b}_a)$, where $elem_a\in\{\mathrm{C,N,O,\dots}\}$.
Each bond is represented as $(u,v,o)\in E$, where $u$ and $v$ are node IDs (atom or super-node) and $o$ is the bond order.
We use string IDs to distinguish node types: atom nodes use the prefix \texttt{A\_} (\eg, \texttt{A\_1}), and functional-group nodes use the prefix \texttt{FG\_} (\eg, \texttt{FG\_1}); IDs are unique within each molecule crop.

\paragraph{Visual Anchors.}
Detecting a functional group thus requires more than bounding boxes: the model must infer which internal attachment sites of a rigid group participate in external bonds. For a group instance $g\in V_{\text{super}}$, we define its \textit{visual anchor set} $\mathcal{A}_g=\{\mathbf{x}^{(1)},\ldots,\mathbf{x}^{(k_g)}\}$, where each anchor $\mathbf{x}^{(j)}=(x,y)$ is the pixel coordinate of an internal atom in $g$ that forms a bond with the external structure (\ie, a residual atom or another super-node). $k_g=|\mathcal{A}_g|$ equals $1$ for terminal groups and $k_g\ge 2$ for bridging/scaffold groups.
In a subsequent step, we crop a group-specific image $I_g$ from the molecule crop using the predicted bounding box $\mathbf{b}_g$, and reuse FG-VLM in a second pass on $I_g$ to output the anchor set $\mathcal{A}_g$.

\paragraph{Serialized Supervision.}
We serialize the hybrid structure prediction (nodes and bonds) as a target sequence $S = S_V \oplus S_E$, where $S_V$ enumerates all nodes and $S_E$ enumerates all bonds:
\begin{equation}
\begin{aligned}
S_V &= \bigoplus_{g\in V_{\text{super}}}\{id_g, name_g, \mathbf{b}_g\} 
 \oplus\ \bigoplus_{a\in V_{\text{atom}}}\{id_a, elem_a, \mathbf{b}_a\}, \\
S_E &= \bigoplus_{(u,v,o)\in E}\{u,v,o\}.
\end{aligned}
\end{equation}

\subsubsection{Structural Reconstruction via Directional Vector Matching.}
\label{ssec:vector_matching}
Given the predicted nodes and bonds $(V,E)$ and the anchor set $\mathcal{A}_g$ for each rigid group $g\in V_{\text{super}}$, reconstructing $G'$ requires orienting $g$ and mapping its anchors to discrete attachment atoms of its class template. A naive alternative is to predict an explicit continuous pose parameter for each group (\eg, a rotation angle) and then rotate the template accordingly. However, such parameterization is brittle under periodicity and rotational symmetries. We instead adopt a purely geometric procedure: \textit{Directional Vector Matching}.

Let $T_{c_g}$ be the canonical 2D template of group class $c_g$ centered at the origin, with $N_{c_g}$ candidate attachment atoms. We denote by $\mathbf{u}_{c_g}^{(i)}\in\mathbb{R}^2$ the template direction vector from the template center to the $i$-th candidate attachment atom ($i\in\{1,\ldots,N_{c_g}\}$). For each predicted anchor $\mathbf{x}^{(j)}\in\mathcal{A}_g$, we compute its direction vector relative to the predicted bounding box center of $g$, where $\mathbf{b}_g=(x_{g,1},y_{g,1},x_{g,2},y_{g,2})$ and $\mathbf{c}_{box}=\big(\frac{x_{g,1}+x_{g,2}}{2},\frac{y_{g,1}+y_{g,2}}{2}\big)$:
\begin{equation}
  \mathbf{v}_{pred}^{(j)} = \mathbf{x}^{(j)} - \mathbf{c}_{box}.
\end{equation}
We map each anchor $j$ to a template index by maximizing cosine similarity:
\begin{equation}
  m^*_j = \arg\max_{i \in \{1,\ldots,N_{c_g}\}} \left( \frac{\mathbf{v}_{pred}^{(j)} \cdot \mathbf{u}_{c_g}^{(i)}}{\| \mathbf{v}_{pred}^{(j)} \| \| \mathbf{u}_{c_g}^{(i)} \|} \right).
\end{equation}
This discretizes rotation estimation into a matching problem. For symmetric groups (\eg, \textit{para}-phenylene), independent matching resolves rotational ambiguity by explicitly mapping each anchor to a specific canonical atom index on the template (\eg, distinguishing the opposing C1 and C4 attachment sites on a benzene ring). Once correspondences are obtained, we align the template to the visual input and instantiate the oriented group in $G'$ by attaching its incident predicted bonds according to the matched anchors.

\subsubsection{Reaction Reconstruction.}
\label{ssec:reconstruction}
Following directional vector matching, we instantiate the complete molecular graph by placing the oriented functional-group templates and attaching the incident predicted bonds to the matched attachment sites, together with the predicted residual atoms. We convert the reconstructed graph into a RDKit\cite{landrum2013rdkit} molecule and export a canonical SMILES string, using RDKit sanitization to validate valence and connectivity. Finally, in the \textit{Reaction Reconstruction} phase, we reassemble the global reaction equation (\eg, Reactants $\to$ Products) by organizing the individual SMILES strings according to the spatial layout parsed in Section~\ref{ssec:deconstruction}, thereby producing a structured reaction representation.

\subsection{Semantic Activation}
\label{ssec:activation}
While FG-VLM provides topology-faithful SMILES and a structured reaction representation, downstream LLMs can still suffer from a semantic disconnect when prompted with SMILES alone. We hypothesize that LLMs index scientific knowledge more effectively through natural-language entity names than through structural strings \cite{zhang2024chemllm}. As a post-recognition enhancement, we introduce a \textit{Semantic Activation} pipeline that converts the recognition output into semantically enriched prompts.

\paragraph{Entity Resolution.}
We employ a deterministic mapping function $\mathcal{M}: \mathcal{Z} \to \mathcal{E}$ that converts the recognized SMILES string $Z$ into a set of entity names $\mathcal{E}$ by querying the corresponding compound record in PubChem \cite{kim2023pubchem}. Specifically, we retrieve the IUPAC name and the synonym list provided by PubChem, and form two name strings: $n_{\text{IUPAC}}$ (the retrieved IUPAC name) and $n_{\text{common}}$ (a comma-separated list of selected synonyms after de-duplication).We use both common names and IUPAC names as they are complementary: common names are frequent in natural-language chemistry texts and better align with LLM pre-training, while IUPAC names provide a standardized, broadly available identifier for long-tail molecules.

\paragraph{Activation Prompting.}
We construct an enriched prompt $P_{act}$ that integrates the visual description with the resolved entities:
\begin{equation}
\begin{aligned}
  P_{act} = &\texttt{"Image shows [} n_{\text{common}} , n_{\text{IUPAC}}\texttt{]} \\
  &\texttt{ (SMILES: [} Z \texttt{]). Analyze..."}
\end{aligned}
\end{equation}
We leverage these name-enhanced prompts to better elicit the model's parametric chemical knowledge, such as reactivity rules, electronic effects, and synthesis priors.

\begin{table*}[t]
\centering
\caption{\textbf{Main Results on OCRD-Bench.} Comparison between the Baseline (Direct Input) and our \OURS{} Framework (Ours). \textbf{Rec}: Recognition, \textbf{Know}: Knowledge, \textbf{Reas}: Reasoning. \OURS{} shows consistent double-digit improvements across all LLMs.}
\label{tab:main_results}
\setlength{\tabcolsep}{5mm}
\resizebox{\textwidth}{!}{%
\begin{tabular}{l|cccc|cccc|c}
\toprule
\multirow{2}{*}{\textbf{Models}} & \multicolumn{4}{c|}{\textbf{Vanilla LLM}} & \multicolumn{4}{c|}{\textbf{LLM+\OURS{}}} & \multirow{2}{*}{\textbf{Gain}} \\
\cmidrule(lr){2-5}\cmidrule(lr){6-9} & Rec & Know & Reas & \textbf{Total} & Rec & Know & Reas & \textbf{Total} & \\
\midrule
\multicolumn{10}{l}{\textit{\textbf{Proprietary SOTA Models}}} \\
\midrule
Gemini 2.5 Pro & 68.0 & 59.0 & 54.0 & 58.80 & 92.0 & 74.0 & 71.0 & \textbf{76.40} & \textcolor{teal}{+17.60} \\
GPT-5 (Preview) & 64.0 & 53.0 & 50.5 & 54.20 & 92.0 & 69.0 & 66.2 & \textbf{72.48} & \textcolor{teal}{+18.28} \\
Qwen3-Max & 61.0 & 50.0 & 48.0 & 51.40 & 92.0 & 70.0 & 65.7 & \textbf{72.68} & \textcolor{teal}{+21.28} \\
GPT-4o & 45.0 & 48.0 & 46.5 & 46.80 & 92.0 & 67.0 & 62.5 & \textbf{70.20} & \textcolor{teal}{+23.40} \\
Claude 4 Sonnet & 49.0 & 47.0 & 45.2 & 46.68 & 92.0 & 66.0 & 62.5 & \textbf{69.80} & \textcolor{teal}{+23.12} \\
\midrule
\multicolumn{10}{l}{\textit{\textbf{Open-Weights}}} \\
\midrule
Yi-Lightning & 47.0 & 41.0 & 38.7 & 41.28 & 92.0 & 58.0 & 54.7 & \textbf{63.48} & \textcolor{teal}{+22.20} \\
Llama-4-109B & 44.0 & 38.5 & 36.7 & 38.88 & 92.0 & 55.0 & 52.0 & \textbf{61.20} & \textcolor{teal}{+22.32} \\
Intern-S1 & 54.0 & 43.5 & 41.2 & 44.68 & 92.0 & 56.5 & 53.5 & \textbf{62.40} & \textcolor{teal}{+17.72} \\
ChemVLM & 36.0 & 27.5 & 25.7 & 28.48 & 92.0 & 32.5 & 30.5 & \textbf{43.60} & \textcolor{teal}{+15.12} \\
\bottomrule
\end{tabular}%
}
\end{table*}

\subsection{Data Construction and Training for FG-VLM}
\label{sec:data_and_training}
This subsection describes how we construct supervision and train FG-VLM to acquire the hybrid-granularity perception and visual-anchor capabilities introduced in Section~\ref{ssec:fgvlm_pipeline}.

\subsubsection{Dataset Construction}
\label{ssec:dataset}
Training FG-VLM requires explicit annotations for functional-group boxes, residual-atom boxes, bond connectivity (node IDs and bond orders), and visual anchors, which are not available in standard atom-level OCSR datasets. To this end, we build \textbf{FG-SFT}, a specialized instruction-tuning dataset for hybrid-granularity recognition. FG-SFT contains 100k molecule samples and 10k reaction samples; detailed statistics and source distributions are provided in Table~\ref{tab:data_stats} (Appendix~\ref{sec:appendix_data}).

\paragraph{Molecule-Anchor Subset.}
We source unique molecular structures from PubChem \cite{kim2023pubchem}. For each molecule, we start from its \texttt{.mol} file and construct the atom--bond graph with bond orders. We then apply a priority-driven greedy decomposition (Appendix~\ref{sec:appendix_data}) to partition atoms into functional-group instances $\{FG_i\}$; atoms not assigned to any $FG_i$ are treated as residual atoms. This yields the hybrid-granularity node set (functional-group nodes + residual-atom nodes) together with the bond set inherited from the \texttt{.mol} connectivity. The atom-wise exclusivity constraint in the decomposition enforces a bijective mapping between each functional-group token and its underlying atom set.
To derive visual anchors, we inspect cross-boundary bonds between the atom set of $FG_i$ and atoms outside $FG_i$ (either residual atoms or atoms in other functional groups). The endpoint atoms \emph{inside} $FG_i$ that participate in such bonds are defined as its attachment atoms and used as anchors. We render each molecule into a 2D diagram with RDKit and obtain anchor keypoints by mapping these attachment atoms to their 2D coordinates in the depiction (converted to pixel coordinates under the drawing canvas), providing the supervision for anchor prediction.

\paragraph{Reaction-Layout Subset.}
We leverage the Open Reaction Database (ORD) \cite{kearnes2021open} to source chemically valid reaction records. To approximate the layouts seen in literature, we synthesize reaction diagrams in three archetypes: \textit{Linear Flow} (70\%), \textit{Multi-line} (15\%), and \textit{Tree/Graph} (15\%); all diagram constituents are automatically annotated with bounding boxes.

\subsubsection{Training Details}
\label{ssec:training}
We fine-tune a single FG-VLM (Qwen2.5-VL-32B)\cite{bai2025qwen2} to perform all perception stages in a unified instruction-following, sequence-generation formulation. Each training example contains a visual input $X$ (a full reaction diagram $I$, a molecule crop $I_{\text{crop}}$, or a functional-group crop $I_g$), a task instruction, and a structured target sequence $Y$ (layout boxes, hybrid nodes+bonds, or anchors).

\paragraph{Supervised Objective.}
We optimize the teacher-forced negative log-likelihood:
\begin{equation}
\mathcal{L}_{\text{NLL}}(X,Y) = - \sum_{t=1}^{|Y|} \log p_\theta\!\left(y_t \mid y_{<t}, X \right),
\end{equation}
where $y_t$ is the $t$-th output token in the serialized target.

\paragraph{Task Definitions.}
We train FG-VLM with three instruction formats:
(i) \textit{Reaction Deconstruction} on $I$: predict diagram constituents and bounding boxes (Section~\ref{ssec:deconstruction});
(ii) \textit{Hybrid Structure Prediction} on $I_{\text{crop}}$: predict the serialized hybrid graph sequence $S=S_V \oplus S_E$ (Section~\ref{ssec:visual_anchor});
(iii) \textit{Anchor Prediction} on $I_g$: predict the anchor list $\mathcal{A}_g=\{\mathbf{x}^{(j)}\}_{j=1}^{k_g}$ for each functional group (Section~\ref{ssec:visual_anchor}).
We denote the corresponding losses as $\mathcal{L}_{\text{layout}}$, $\mathcal{L}_{\text{graph}}$, and $\mathcal{L}_{\text{anchor}}$.

\paragraph{Three-Stage Curriculum.}
We adopt a curriculum-style schedule that progressively introduces finer-grained supervision:
\begin{itemize}
    \item Stage I (Graph pretraining). We first train FG-VLM on the molecule subset (100k molecules) to learn hybrid structure prediction on $I_{\text{crop}}$, optimizing $\mathcal{L}_{\text{graph}}$ only.
    \item Stage II (Anchor refinement). We then add anchor supervision on functional-group crops $I_g$ from the same molecules, jointly optimizing $\mathcal{L}_{\text{graph}}+\mathcal{L}_{\text{anchor}}$ to improve fine-grained attachment perception while retaining hybrid graph prediction ability.
    \item Stage III (Joint instruction tuning). Finally, we mix reaction-layout samples (10k reactions) with the molecule-level graph and anchor samples, and optimize a weighted multi-task objective:
    \begin{equation}
        \begin{aligned}
        \mathcal{L} \;=\;&
        \lambda_{\text{layout}}\mathcal{L}_{\text{layout}}
        + \lambda_{\text{graph}}\mathcal{L}_{\text{graph}} + \lambda_{\text{anchor}}\mathcal{L}_{\text{anchor}}.
        \end{aligned}
    \end{equation}
\end{itemize}

In Stage III, mini-batches are drawn from the three task pools with fixed sampling probabilities (treated as hyperparameters), while all tasks share the same FG-VLM parameters across input granularities.

\paragraph{Cropping Strategy.}
During training, molecule crops $I_{\text{crop}}$ and group crops $I_g$ are generated from ground-truth bounding boxes to avoid compounding errors; at inference, they are generated from FG-VLM predictions.

\section{Experiments}

\subsection{Experimental Setup}
\label{sec:setup}

\paragraph{Implementation Details.}
For the Hybrid-Granularity Visual Module, we use Qwen2.5-VL-32B as the perception backbone and fine-tune it on our FG-SFT dataset to support functional-group grounding and visual anchor detection.
For the Semantic Activation stage, we query the evaluation LLMs with structured outputs (SMILES and entity names) using a standardized prompt that enforces a ``Name-First'' reasoning trajectory: ``\textit{Identify the reaction type based on the provided chemical name, then analyze the mechanism...}''.
All hyperparameters (\eg, learning rates, batch sizes) and the complete prompt templates are provided in Appendix.

\paragraph{Baselines.}
We benchmark \OURS{} against a comprehensive suite of multimodal LLMs, grouped into proprietary SOTA models and open-weights models (Table \ref{tab:main_results}). In the \textit{Direct Input} baseline, we supply the raw reaction image to each model and rely on its native visual encoder, which serves as a lower bound for end-to-end performance.

\paragraph{Evaluation Dataset.}
To rigorously evaluate diagram-grounded reasoning, we introduce \textbf{OCRD-Bench}, a challenging benchmark curated from graduate-level chemistry assessment materials. Unlike prior datasets centered on either isolated structure recognition or multiple-choice answering, OCRD-Bench emphasizes deep, multi-step logical deduction.
Specifically, OCRD-Bench contains 100 core reaction scenarios expanded into 500 hierarchical questions, spanning eight major categories that cover the pillars of organic chemistry (see the taxonomy in Appendix Figure \ref{fig:taxonomy}). We organize evaluation into three cognitive tiers that mirror the scientific reasoning pipeline: L1 (Visual Perception) measures grounding of visual evidence into symbolic representations; L2 (Chemical Knowledge) assesses retrieval of static reactivity rules and reaction conditions; and L3 (Mechanistic Reasoning) requires generating step-by-step electron-flow explanations and key intermediates (Figure \ref{fig:example}). To ensure high-quality ground truth, two PhD-level domain experts independently verified and corrected all standard keys to
ensure scientific accuracy and support reliable automatic evaluation.

\paragraph{Evaluation Metric.}
We employ task-specific metrics to enable a rigorous assessment. For L1, we report strict binary accuracy based on the Tanimoto similarity of Morgan fingerprints. For L2 and L3, accuracy alone is insufficient: L2 uses multiple-choice questions where the \textit{rationale} is critical, whereas L3 is open-ended.
We therefore adopt the LLM-as-a-Judge paradigm \cite{zheng2023judging}, using Gemini 2.5 Pro to assess both option correctness and the logical validity of the accompanying explanation against the reference answers. Formal metric definitions and judge prompts are provided in Appendix \ref{sec:appendix_benchmark_metrics}.


\begin{table*}[t]
\centering
\caption{Scale Robustness on Qwen2.5-VL. Vnilla LLM vs.\ \OURS{} across different parameter scales within the same model family. \OURS{} yields stable gains, indicating low sensitivity to backbone size.}
\label{tab:qwen_scaling}
\setlength{\tabcolsep}{5mm}
\resizebox{\textwidth}{!}{%
\begin{tabular}{l|cccc|cccc|c}
\toprule
\multirow{2}{*}{\textbf{Backbone}} & \multicolumn{4}{c|}{\textbf{Vanilla LLM}} & \multicolumn{4}{c|}{\textbf{LLM+\OURS{}}} & \multirow{2}{*}{\textbf{Gain}} \\
\cmidrule(lr){2-5}\cmidrule(lr){6-9}
 & Rec & Know & Reas & \textbf{Total} & Rec & Know & Reas & \textbf{Total} & \\
\midrule
Qwen2.5-VL-3B  & 11.0 & 9.0  & 3.1  & 6.42  & 92.0 & 23.0 & 11.6 & 32.24 & \textcolor{teal}{+25.82} \\
Qwen2.5-VL-7B  & 21.0 & 18.5 & 13.5 & 14.30 & 92.0 & 35.0 & 22.1 & 41.24 & \textcolor{teal}{+26.94} \\
Qwen2.5-VL-32B & 39.0 & 28.5 & 24.7 & 24.14 & 92.0 & 44.5 & 39.7 & 52.08 & \textcolor{teal}{+27.94} \\
Qwen2.5-VL-72B & 45.0 & 36.0 & 32.2 & 29.84 & 92.0 & 51.0 & 45.6 & 57.04 & \textcolor{teal}{+27.20} \\
\bottomrule
\end{tabular}%
}
\end{table*}

\subsection{Main Results}
\label{sec:main_results}

Table \ref{tab:main_results} compares the performance of the proposed \OURS{} framework with that of the Vanilla LLM baseline.

\paragraph{Overcoming the Visual Deficit.}
The baseline results strongly validate the ``Visual Deficit'' hypothesis. Even leading proprietary models struggle to interpret the topological semantics of molecular graphs. 
While Gemini-2.5-Pro achieves a recognition score of 68\%, other large-scale models such as GPT-4o perform poorly, attaining only 45\% accuracy in structural identification via direct input. This observation suggests that general-purpose vision encoders constitute the primary bottleneck. Specialized models such as ChemVLM also exhibit limitations, with a Recognition score of 36\%, which restricts their effectiveness in this domain. In contrast, our \OURS{} framework explicitly models functional groups and anchors, elevating the Recognition score to {92\%} across all backbones. This structural correction provides a consistent and accurate foundation for downstream tasks.

\paragraph{Performance Leap in Knowledge and Reasoning.}
Beyond addressing visual perception, the proposed framework achieves substantial gains in downstream reasoning tasks (refer to the Know \& Reas columns in Table \ref{tab:main_results}). 
Our \OURS{} framework significantly enhances the Total Score across all architectures. This improvement is particularly pronounced for Open-Weights models: the Total Score of Llama-4-109B increases from 38.88\% to 61.2\% (+22.32), while that of Yi-Lightning improves from 41.28\% to 63.48\% (+22.2). 
Crucially, this comprehensive improvement enables open-weights models to compete with proprietary SOTA systems. With the proposed framework, Llama-4-109B (61.2\%) outperforms the direct input baseline of Gemini 2.5 Pro (58.8\%). This empirically demonstrates that providing structured, high-quality inputs that incorporate both precise visual graphs and semantic entities is critical for realizing the reasoning potential of these models.

\subsection{Robustness to Backbone Scale}
\label{sec:scale}


While Table~\ref{tab:main_results} compares a diverse set of architectures, variations across model families may obscure the specific effects of parameter scaling.
To isolate the impact of backbone size, we conduct a controlled scaling study using the Qwen2.5-VL family, with parameter counts ranging from 3B to 72B.
Each model is evaluated under two configurations: (i) \textit{Vanilla LLM}, in which the model generates responses directly from raw reaction images via its native vision encoder; and (ii) \textit{LLM+\OURS{}}, in which the model relies on topology-faithful structures and entity names generated by FG-VLM and the Semantic Activation pipeline.

As presented in Table~\ref{tab:qwen_scaling}, \OURS{} achieves a consistent absolute improvement ranging from +25.82 to +27.94 points in the Total score across all model sizes.
Although the performance of the Direct Input baseline improves with scale (increasing from 6.42 to 29.84), it remains significantly constrained by visual perception capabilities (with Rec $\leq$ 45\% even at the 72B scale).
Conversely, \OURS{} elevates the recognition score to 92.0\% for all backbones and consistently enhances performance in both the Knowledge and Mechanistic Reasoning tasks.
These results indicate that the primary bottleneck is unreliable visual grounding rather than insufficient linguistic capacity. Consequently, employing topology-faithful recognition combined with semantic activation offers an efficient alternative to simply increasing parameter scale.

\subsection{Ablation Studies}
\label{sec:ablation}

To rigorously validate the effectiveness of our framework components, we conduct ablation studies focusing on two critical aspects: the impact of incorporating semantic entities (Semantic Activation) and the necessity of high-fidelity structure recognition (FG-VLM).

\begin{figure}[H]
    \centering
    \includegraphics[width=1.0\linewidth]{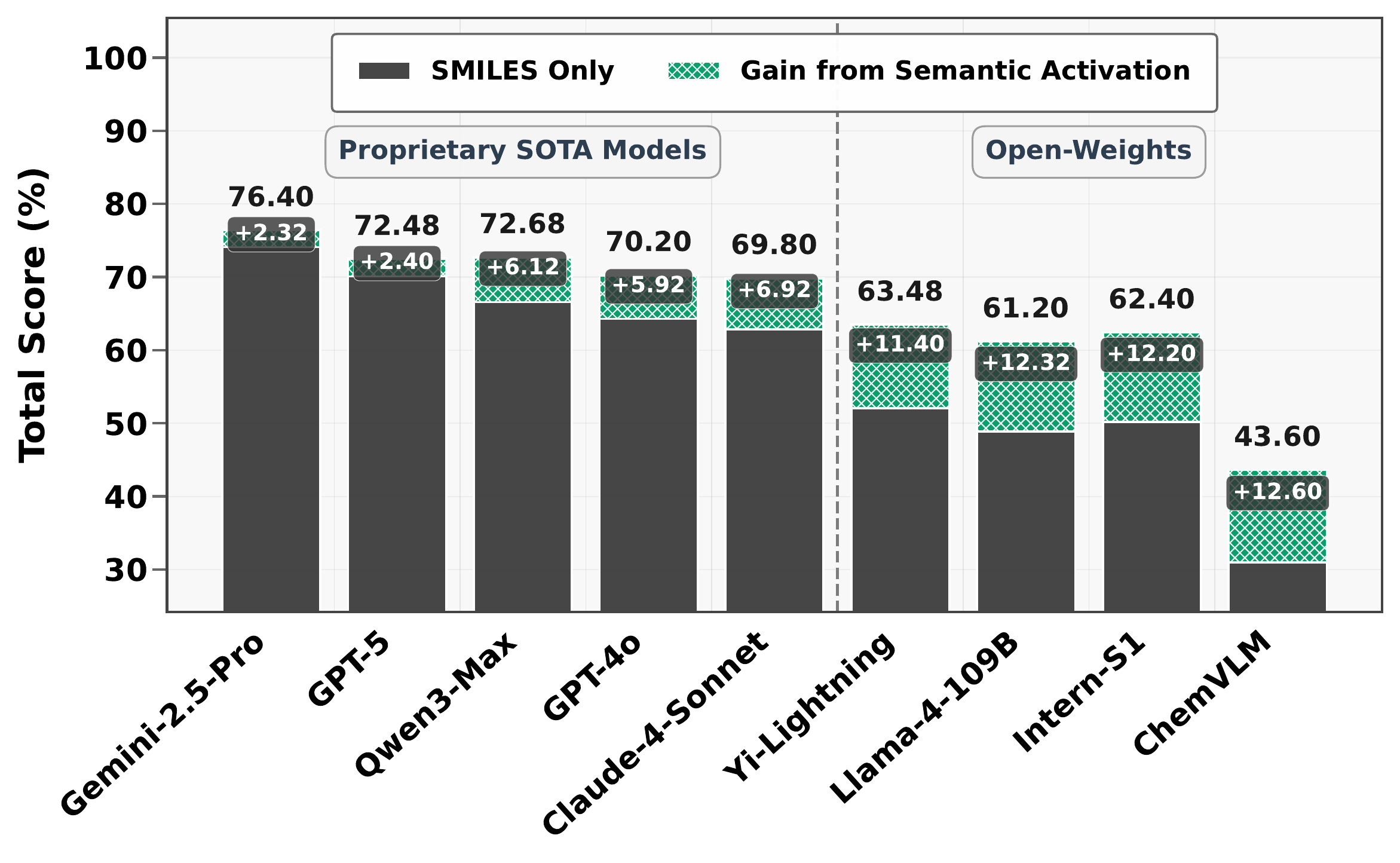}
    \caption{Effectiveness of Semantic Activation. The stacked bar chart illustrates the Total Scores across all models. The gray base represents the performance with \textit{SMILES Only}, while the green top section highlights the Gain from Semantic Activation. The contrast clearly shows that gains are universal, with all models demonstrating significant improvement.}
    \label{fig:ablation_chart}
\end{figure}

\paragraph{Bridging the Semantic Disconnect.}
For qualitative case studies and detailed performance metrics, please refer to Appendix \ref{app:detailed_data}.
As illustrated in Figure \ref{fig:ablation_chart}, the results demonstrate that \textit{Semantic Activation} serves as an effective solution to the semantic disconnect.
By incorporating Entity Names, consistent performance gains are observed across the entire spectrum of models.
This effect is particularly substantial in Open-Weights models, where improvements exceed 10\%. For instance, the score of Llama-4-109B improves from a baseline of 48.88\% (see Appendix Table \ref{tab:ablation_breakdown}) to 61.2\% (+12.3\%), and that of Yi-Lightning increases from 52.08\% to 63.5\% (+11.4\%).
This consistent enhancement confirms that \OURS{} effectively bridges the semantic gap, enabling diverse architectures to accurately align visual representations with latent chemical reasoning.

\begin{figure}[H]
    \centering
    \includegraphics[width=1.0\linewidth]{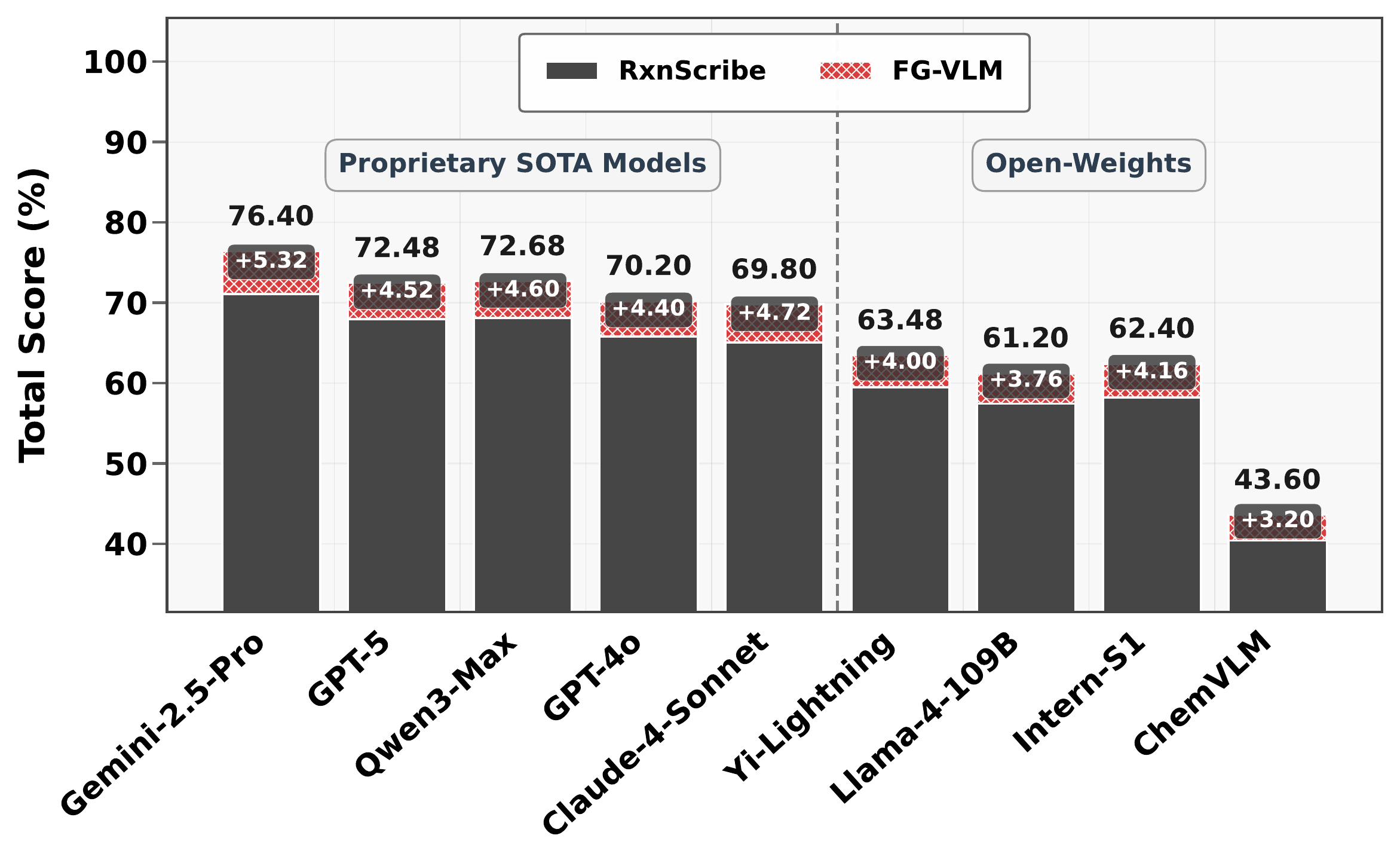}
    \caption{Superiority of FG-VLM over RxnScribe. The chart illustrates the Total Score comparison across 9 models. The gray bars represent the performance using RxnScribe as the recognition tool, while the red sections indicate the explicit performance gains achieved by our FG-VLM. }
    \label{fig:ablationstudy_molscribe}
\end{figure}

\paragraph{Replacing the Recognition Frontend.}
To further verify that the observed gains are driven by accurate structure recognition, the FG-VLM module is replaced with a range of off-the-shelf recognizers. These include transformer-based OCSR tools (\eg, RxnScribe~\cite{qian2023rxnscribe} and DECIMER~\cite{rajan2020decimer,decimer2023}) and instruction-tuned chemical VLMs (\eg, ChemVLM~\cite{li2024chemvlm}).
Table~\ref{tab:ocsr_swap} presents the L1 recognition accuracies on OCRD-Bench.
While transformer-based OCSR baselines reach 83--86\%, VLM-based recognizers exhibit severe hallucinations (25--36\%), highlighting the limitations of generic encoders.
In contrast, the proposed FG-VLM achieves {92\%}, surpassing all alternatives.
Furthermore, the most competitive external baseline (RxnScribe) is integrated into the pipeline to evaluate end-to-end performance across all nine LLM backbones.
As depicted in Figure~\ref{fig:ablationstudy_molscribe} and Appendix Table~\ref{tab:molscribe_data}, the recognition ceiling of RxnScribe (86\%) constrains the final Total Scores (\eg, Llama-4-109B: 57.44 with RxnScribe vs.\ 61.20 with FG-VLM), demonstrating that topology-faithful recognition is a prerequisite for reliable downstream reasoning.

\begin{table}[t]
\centering
\caption{\textbf{Recognizer Substitution on OCRD-Bench (L1).} We compare several structure recognizers as drop-in replacements for FG-VLM.}
\label{tab:ocsr_swap}
\setlength{\tabcolsep}{3mm}
\resizebox{\columnwidth}{!}{%
\begin{tabular}{l l c}
\toprule
\textbf{Model Type} & \textbf{Model Name} & \textbf{OCRD-Bench (L1 only)} \\ 
\midrule
\multirow{2}{*}{Transformer} & RxnScribe & 86\% \\
 & Decimer & 83\% \\ 
\midrule
\multirow{2}{*}{VLM} & ChemDFM-X-13B & 25\% \\
 & ChemVLM-26B & 36\% \\
\midrule
\textbf{Ours} & \textbf{FG-VLM} & \textbf{92\%} \\ 
\bottomrule
\end{tabular}%
}
\end{table}


\section{Conclusion}

In this work, we present a unified framework for chemical visual reasoning that bridges the critical gaps of visual deficit and semantic disconnect via hybrid-granularity anchoring and semantic activation. Built upon this framework, we develop a specialized visual module (FG-VLM) and a semantic alignment pipeline, capable of precise functional group grounding and effective activation of latent chemical knowledge in Multimodal LLMs. Moreover, we contribute two high-quality resources to the community: FG-SFT, a large-scale hybrid-granularity dataset for fine-grained visual instruction tuning, and OCRD-Bench, a challenging evaluation benchmark derived from graduate-level exams requiring end-to-end mechanistic reasoning. Extensive experiments have demonstrated that \OURS{} acts as a powerful performance equalizer, enabling open-weights models to surpass proprietary SOTA systems and achieving superior structural recognition accuracy compared to specialized tools.
\section*{Limitations and Ethical Considerations }
\label{sec:limitations}

While our \OURS{} framework improves chemical reasoning, two limitations remain:
The \textit{Visual Anchor Mechanism} relies on a predefined dictionary of functional groups. While our current vocabulary covers over 95\% of common substructures (\eg, in PubChem), the model must revert to atom-level prediction for rare or atypical groups outside this set. This fallback loses the geometric robustness provided by the anchor vectors, potentially reducing recognition accuracy for exotic molecules.
In addition, the \textit{Semantic Alignment} module depends on mapping recognized structures (SMILES) to meaningful entity names. For novel molecules or intermediates that lack established common names or database entries, the "Knowledge Activation" pipeline cannot trigger the LLM's latent semantic networks.
In addition, this work focuses on chemical reaction diagram understanding for research purposes, and does not involve human subjects, personal data, or deployment in safety-critical settings; we therefore do not anticipate specific ethical concerns beyond standard dataset licensing and responsible use of models.

\clearpage
\bibliographystyle{ACM-Reference-Format}
\bibliography{ref}

\clearpage

\appendix\label{sec:appendix}
\setcounter{table}{0}   
\setcounter{figure}{0}
\setcounter{section}{0}
\setcounter{equation}{0}
\renewcommand{\thetable}{A\arabic{table}}
\renewcommand{\thefigure}{A\arabic{figure}}
\renewcommand{\thesection}{A\arabic{section}}
\renewcommand{\theequation}{A\arabic{equation}}

\section{Implementation Details: Prompts and Instruction Tuning}
\label{sec:appendix_prompts}
\subsection{Reaction Diagram Deconstruction Prompt}
\label{ssec:prompt_parsing}

To initiate the pipeline, we employ the following prompt to decompose the raw reaction image into semantic components. Beyond simple object detection, this prompt enforces a \textbf{logic-aware grouping mechanism} via the \texttt{reaction\_id} field. This design is critical for handling complex scenarios such as multi-step synthesis (\eg, $A \to B \to C$) or disconnected reaction variants, ensuring that shared intermediates are correctly associated with their respective reaction steps. The structured JSON output containing bounding boxes is subsequently used to crop individual molecular images for fine-grained recognition.

\begin{tcolorbox}[colback=blue!3, colframe=black, arc=2mm, left=3pt, right=3pt,
    boxsep=5pt, boxrule=0.5pt, colframe=black!70, fonttitle=\bfseries, title = Reaction Parsing Prompt, breakable]

\textbf{System Prompt} \\
You are an expert chemical reaction diagram analysis assistant. Your task is to analyze chemical reaction diagrams and parse them into structured component lists.
You must accurately detect all visual elements, classify their roles, and group them by reaction pathways.
Handle complex layouts including linear flows, multi-step synthesis, and divergent/convergent pathways. Respond strictly in the specified JSON format. \\

\textbf{User Prompt} \\
\#\#\# Task Description \#\#\# \\
Analyze the provided chemical reaction diagram image and extract all semantic components.
For each detected object, provide:
1. \textbf{Reaction ID}: An integer identifier grouping components belonging to the same reaction step. If multiple distinct reactions are present (\eg, Step 1 and Step 2), assign distinct IDs. Shared components (\eg, an intermediate serving as product of R1 and reactant of R2) should be listed for \textit{each} reaction they participate in.
2. \textbf{Role}: Classify the object into one of: "Reactant", "Product", "Condition", "Arrow", "Plus".
3. \textbf{BBox}: The bounding box coordinates [x1, y1, x2, y2] normalized to 0-1000.

\#\#\# Parsing Rules \#\#\# \\
\textbf{Role Definitions}: \\
- \textit{Reactant}: Starting materials or substrates appearing before the reaction arrow. \\
- \textit{Product}: Final compounds or intermediates appearing after the reaction arrow. \\
- \textit{Condition}: Text or small formulas above/below arrows (reagents, solvents, temperature, time). \\
- \textit{Arrow}: The graphical arrow indicating reaction direction. \\
- \textit{Plus}: The "+" sign separating multiple reactants or products. \\

\textbf{Grouping Logic}: \\
- A "Reaction" consists of a set of Reactants + Conditions + Arrow $\to$ Products.
- In multi-step reactions ($A \to B \to C$), "B" is the Product of Reaction 1 and the Reactant of Reaction 2.
- In disconnected layouts (multiple independent reactions in one image), treat them as separate Reaction IDs.

\#\#\# Output Format (STRICT) \#\#\# \\
Respond with a valid JSON list of objects. Do not include markdown code blocks (```json).
[
  \{"reaction\_id": 1, "role": "Reactant", "bbox": [x1, y1, x2, y2]\},
  \{"reaction\_id": 1, "role": "Arrow", "bbox": [x1, y1, x2, y2]\},
  \{"reaction\_id": 1, "role": "Condition", "bbox": [x1, y1, x2, y2]\},
  \{"reaction\_id": 1, "role": "Product", "bbox": [x1, y1, x2, y2]\},
  \{"reaction\_id": 2, "role": "Reactant", "bbox": [x1, y1, x2, y2]\},
  ...
]

\#\#\# Input Image \#\#\# \\
\{input\_image\} \\

Now begin your structured response:

\end{tcolorbox}

\subsection{Hybrid-Granularity Structure Recognition Prompt}
\label{ssec:prompt_recognition}

This prompt drives the core hybrid-granularity recognition. The prompt enforces a \textbf{Top-Down Visual Perception} strategy. We instruct the model to prioritize the detection of holistic functional group patterns based on the provided Priority List , treating atoms merely as residual elements to fill the gaps. This "Chunking" strategy ensures topological robustness against visual noise.

\begin{tcolorbox}[colback=blue!3, colframe=black, arc=2mm, left=3pt, right=3pt,
    boxsep=5pt, boxrule=0.5pt, colframe=black!70, fonttitle=\bfseries, title = Hybrid-Granularity Structure Recognition, breakable]

\textbf{System Prompt} \\
You are an expert chemical structure recognition assistant. Your task is to convert a cropped molecule image into a \textbf{Hybrid-Granularity Graph}.
You must employ a \textbf{Top-Down Visual Perception} strategy:
1. First, identify large, rigid \textbf{Super-nodes} (Functional Groups) based on global visual patterns.
2. Second, identify \textbf{Isolated Atoms} only in the remaining empty regions.
3. Finally, determine the connectivity (\textbf{Bonds}) between all identified nodes.

\textbf{User Prompt} \\
\#\#\# Task Description \#\#\# \\
Analyze the molecular image and generate a structured JSON representation containing \texttt{supernodes}, \texttt{atoms}, and \texttt{bonds}.

\#\#\# Decomposition Constraints (CRITICAL) \#\#\# \\
1. \textbf{Visual Priority (Top-Down)}:
   - \textit{Rule}: Prioritize the detection of Functional Group Patterns (from \{priority\_list\_file\}) over individual atoms.
   - \textit{Strategy}: Treat the image primarily as a composition of Functional Groups. Identify these rigid structures based on their global visual features.
   - \textit{Conflict Resolution}: If a visual region matches a Functional Group template, you MUST output it as a Super-node. Do NOT decompose it into internal atoms.
   
2. \textbf{Reference Priority List}: Strict adherence to the provided file: \textbf{\{priority\_list\_file\}}.
   - \textit{Greedy Matching}: Scan and match functional groups strictly in the order they appear in the file.
   - \textit{Closed-Set Constraint}: Only recognize Super-nodes explicitly listed in \{priority\_list\_file\}. If a substructure is not in the list, treat it as distinct Isolated Atoms.

3. \textbf{Residual Atom Identification}:
   - \textit{Rule}: Only scan for \textbf{Isolated Atoms} in the image regions that are \textbf{NOT} covered by the bounding boxes of identified Super-nodes.
   - \textit{Logic}: Atoms are "residual" elements. If an atom is visually encapsulated within a Super-node's region, it is implicitly handled by the Super-node template and must not be listed here.
   
4. \textbf{Super-node Exclusivity (No Overlap)}:
   - \textit{Rule}: Super-nodes must NOT share any common atoms.
   - \textit{Logic}: While bounding boxes of adjacent groups may spatially overlap due to crowding, the underlying atomic membership is mutually exclusive. No single atom can belong to more than one Super-node. If ambiguity exists, assign ownership to the group appearing earlier in the Priority List.
\#\#\# Output Schema Definitions \#\#\# \\
1. \textbf{SuperNodes}:
   - \texttt{id}: Unique identifier (\eg, "fg\_1").
   - \texttt{label}: Must match a name from \{priority\_list\_file\}.
   - \texttt{bbox}: [x1, y1, x2, y2].

2. \textbf{Atoms} (Residual):
   - \texttt{id}: Unique identifier (\eg, "atom\_1").
   - \texttt{symbol}: Element symbol (\eg, "C", "N", "O").
   - \texttt{bbox}: [x1, y1, x2, y2]. 

3. \textbf{Bonds}:
   - \texttt{source}: Node ID (Atom or SuperNode).
   - \texttt{target}: Node ID (Atom or SuperNode).
   - \texttt{type}: Bond order ("SINGLE", "DOUBLE", "TRIPLE", "AROMATIC", "WEDGE", "DASH"). 
   *Note: Connects any pair of nodes (Atom-Atom, SuperNode-SuperNode, or Atom-SuperNode).*

\#\#\# Output Format (STRICT) \#\#\# \\
Respond with a single valid JSON object. \\
\texttt{\{} \\
\texttt{  "supernodes": [ ... ],} \\
\texttt{  "atoms": [ ... ],} \\
\texttt{  "bonds": [ ... ]} \\
\texttt{\}}

\#\#\# Input Image \#\#\# \\
\{cropped\_molecule\_image\} \\

\#\#\# Functional Group Priority List (File X) \#\#\# \\
\{priority\_list\_content\} \\

Now begin your analysis, prioritizing visual patterns from the list:

\end{tcolorbox}

\subsection{Fine-Grained Visual Anchor Prediction Prompt}
\label{ssec:prompt_anchor}

While Phase 2 establishes the topological connectivity, it lacks the precise geometric alignment required for reconstructing the chemical drawing. To bridge this gap, Phase 3 employs a specialized \textbf{Visual Grounding Prompt}. This prompt takes the bounding box and class label of a detected super-node (from Phase 2) as input and queries the VLM for the precise coordinates of its \textbf{Visual Anchors}—the specific atoms within the group that serve as attachment points to the external skeleton. These coordinates are the prerequisites for the subsequent \textit{Directional Vector Matching} algorithm.

\begin{tcolorbox}[colback=blue!3, colframe=black, arc=2mm, left=3pt, right=3pt,
    boxsep=5pt, boxrule=0.5pt, colframe=black!70, fonttitle=\bfseries, title = Prompt for Phase 3: Visual Anchor Prediction, breakable]

\textbf{System Prompt} \\
You are a fine-grained visual grounding assistant for chemical structures.
Your task is to identify the precise "Visual Anchors" within a specific functional group.
\textbf{Definition}: A Visual Anchor is an atom \textit{belonging to the functional group} that forms a chemical bond with an atom \textit{outside} the group.
- \textbf{Terminal Groups} (\eg, -OH, -Cl): Typically have 1 anchor.
- \textbf{Linker Groups} (\eg, Phenyl, Pyridine): Can have 2 or more anchors depending on the substitution pattern (para-, meta-, ortho-).

\textbf{User Prompt} \\
\#\#\# Task Description \#\#\# \\
Focus exclusively on the functional group located at the bounding box: \textbf{\{target\_bbox\}}.
The class of this group is: \textbf{\{target\_label\}}.

Your goal is to find the attachment points to the external molecule skeleton.
1. Observe the specified region.
2. Identify the specific atom(s) inside this group that connect to external bonds.
3. Return the center coordinates [x, y] of these anchor atoms (normalized 0-1000).

\#\#\# Output Format (STRICT) \#\#\# \\
Respond with a JSON object containing a single list of coordinates. \\
\texttt{\{} \\
\texttt{  "anchors": [} \\
\texttt{    [x1, y1],} \\
\texttt{    [x2, y2]} \\
\texttt{  ]} \\
\texttt{\}}

\#\#\# Input Image \#\#\# \\
\{cropped\_molecule\_image\} \\

Now, locate the anchors for the \textbf{\{target\_label\}} at \textbf{\{target\_bbox\}}:

\end{tcolorbox}


\section{Data Construction Details}
\label{sec:appendix_data}

\subsection{Image Synthesis Configuration}

Standard datasets typically offer binary modalities: either high-level text captions (\eg, molecule names) or low-level dense coordinates (\eg, pure ``.mol'' files). Neither format is sufficient to teach a model the expert-like cognitive process of identifying functional groups.

To address this, we constructed \textbf{FG-SFT}, a specialized dataset designed to {instill hybrid-granularity perception} into the model. Unlike simple Image-SMILES pairs, FG-SFT provides a structured annotation layer that sits between raw atoms and semantic entities. It comprises {100k} Molecule samples and 10k Reaction samples, where each entry explicitly maps visual regions to functional group  and defines their specific \textit{attachment points} (anchors). This unique format compels the model to learn both the semantic identity of chemical substructures and their precise geometric connectivity within the global reaction context.
To ensure visual robustness, we synthesized images using 8 distinct templates, varying parameters such as bond thickness, font style, and atom label visibility. For reaction diagrams, we leveraged the Open Reaction Database (ORD) \cite{kearnes2021open}.
\begin{table}[h]
\centering
\caption{\textbf{Statistics of the FG-SFT Training Dataset.} The dataset is balanced across molecular recognition and reaction reasoning tasks.}
\setlength{\tabcolsep}{5mm}
\label{tab:data_stats}
\resizebox{\linewidth}{!}{
\begin{tabular}{l|c|c}
\toprule
\textbf{Data Subset} & \textbf{Source} & \textbf{Sample Count} \\
\midrule
Reaction-Linear & $\dagger$ & 7,000 \\
Reaction-MultiLine & $\dagger$ & 1,500 \\
Reaction-Graph & $\dagger$ & 1,500 \\
Molecule & $\ddagger$ & 100,000 \\
\bottomrule
\end{tabular}
}
\begin{minipage}{0.9\linewidth} 
    \footnotesize 
    \vspace{0.3em}
    \textit{Note:} $\dagger$: Sourced from ORD/Open-Source Datasets. $\ddagger$: Sourced from PubChem/Open-Source Datasets.
\end{minipage}
\end{table}


\lstdefinelanguage{json}{%
  basicstyle=\ttfamily\footnotesize,
  showstringspaces=false,
  breaklines=true,
  literate=
   *{0}{{{0}}}{1}{1}{{{1}}}{1}{2}{{{2}}}{1}{3}{{{3}}}{1}{4}{{{4}}}{1}
    {5}{{{5}}}{1}{6}{{{6}}}{1}{7}{{{7}}}{1}{8}{{{8}}}{1}{9}{{{9}}}{1}
    {:}{{{:}}}{1}{,}{{{,}}}{1}{\{}{{{\{}}}{1}{\}}{{{\}}}}{1}
    {[}{{{[}}}{1}{]}{{{]}}}{1},
}


\lstdefinelanguage{json}{%
  basicstyle=\ttfamily\footnotesize,
  showstringspaces=false,
  breaklines=true,
  literate=
   *{0}{{{0}}}{1}{1}{{{1}}}{1}{2}{{{2}}}{1}{3}{{{3}}}{1}{4}{{{4}}}{1}
    {5}{{{5}}}{1}{6}{{{6}}}{1}{7}{{{7}}}{1}{8}{{{8}}}{1}{9}{{{9}}}{1}
    {:}{{{:}}}{1}{,}{{{,}}}{1}{\{}{{{\{}}}{1}{\}}{{{\}}}}{1}
    {[}{{{[}}}{1}{]}{{{]}}}{1},
}

\begin{figure}[!htp]
  \centering
  \includegraphics[width=0.95\columnwidth]{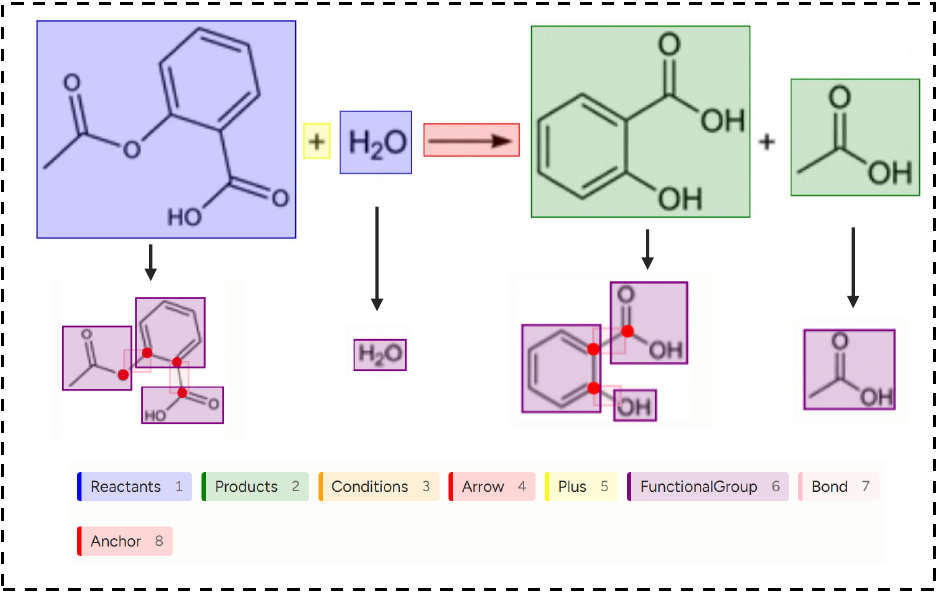}
  \caption{\textbf{FG-SFT example.} A raw reaction diagram with our FG-SFT annotations, including component-level boxes (Reactants, Products, Conditions, Arrow, Plus) and fine-grained labels (functional-group boxes, bond boxes, and anchor keypoints).}
  \label{FG-sft}
\end{figure}

\begin{tcolorbox}[
  colback=blue!3,
  colframe=black!70,
  arc=2mm,
  left=3pt,right=3pt,
  boxsep=5pt,
  boxrule=0.5pt,
  fonttitle=\bfseries,
  title=FG-SFT COCO-style Annotation\cite{lin2014microsoft} (Example),
  width=\linewidth,
  breakable
]
\begin{lstlisting}[
  language=json,
  basicstyle=\ttfamily\footnotesize,
  breaklines=true,
  breakatwhitespace=false,
  columns=fullflexible
]
{
  "images": [
    { "id": 0, "width": 438, "height": 149 },
    { "id": 1, "width": 94,  "height": 94  },
    { "id": 2, "width": 26,  "height": 22  },
    { "id": 3, "width": 82,  "height": 66  },
    { "id": 4, "width": 45,  "height": 43  }
  ],
  "categories": [
    { "id": 0, "name": "Arrow" },
    { "id": 1, "name": "Conditions" },
    { "id": 2, "name": "FunctionalGroup" },
    { "id": 3, "name": "Plus" },
    { "id": 4, "name": "Products" },
    { "id": 5, "name": "Reactants" },
    { "id": 6, "name": "Bond" },
    { "id": 7, "name": "Anchor" }
  ],
  "annotations": [
    { "id": 0, "image_id": 0, "category_id": 5, "bbox": [9.842696426713314, 24.11469943321664, 123.0337081689187, 102.85602809804595] },
    { "id": 1, "image_id": 0, "category_id": 5, "bbox": [154.03820224719104, 66.93033707865169, 33.95730337078655, 29.528089887640455] },
    { "id": 2, "image_id": 0, "category_id": 4, "bbox": [245.0831458649156, 26.575280898876393, 103.84044964070257, 90.55278370922488] },
    { "id": 3, "image_id": 0, "category_id": 0, "bbox": [193.90112359550562, 72.83595505617978, 45.276404494382035, 15.256179775280895] },
    { "id": 4, "image_id": 0, "category_id": 4, "bbox": [368.6089887640451, 51.67419100942299, 61.024719101123544, 55.119067417543256] },
    { "id": 5, "image_id": 0, "category_id": 3, "bbox": [136.81348314606745, 72.83595505617978, 12.303370786516838, 16.240449438202244] },

    { "id": 6, "image_id": 1, "category_id": 2, "bbox": [42.772914430818844, 11.654662055488608, 38.71857656274783, 39.11501050858761] },
    { "id": 7, "image_id": 1, "category_id": 2, "bbox": [1.617676919769829, 27.48886108930232, 38.586436511857926, 36.339972887709216] },
    { "id": 8, "image_id": 1, "category_id": 2, "bbox": [46.158279007543406, 62.07354685827545, 45.59014143850223, 20.218239461161858] },
    { "id": 9, "image_id": 1, "category_id": 6, "bbox": [36.22133942935225, 41.036336247198605, 14.663592154970253, 12.256130208981267] },
    { "id": 10, "image_id": 1, "category_id": 6, "bbox": [61.60886594821314, 47.71122806061365, 10.942943747285096, 17.946525800088715] },
    { "id": 11, "image_id": 1, "category_id": 7, "bbox": [69, 65, 0, 0], "keypoints": [69, 65, 2] },

    { "id": 12, "image_id": 2, "category_id": 2, "bbox": [1.4982595024672898, 4.630889830131542, 24.50174049753269, 14.664484462083207] },

    { "id": 13, "image_id": 3, "category_id": 2, "bbox": [43.831353712647655, 1.6233789731164145, 35.714393151953395, 37.74365447296713] },
    { "id": 14, "image_id": 3, "category_id": 2, "bbox": [45.454723666369084, 51.94825561870752, 19.48059585701531, 14.051744381292487] },
    { "id": 15, "image_id": 3, "category_id": 2, "bbox": [2.4350940336716946, 21.509836637025675, 36.52609768035893, 40.58456165346862] },
    { "id": 16, "image_id": 3, "category_id": 6, "bbox": [35.53113426627757, 23.018826070221383, 14.995638526775984, 11.748175672830737] },
    { "id": 17, "image_id": 3, "category_id": 6, "bbox": [35.91312944139563, 49.95370849328054, 12.51232708738311, 9.073811275111346] },
    { "id": 18, "image_id": 3, "category_id": 7, "bbox": [52, 24, 0, 0], "keypoints": [52, 24, 2] },

    { "id": 19, "image_id": 4, "category_id": 2, "bbox": [3.0908032596041917, 2.409778812572759, 41.90919674039581, 36.69455735869402] }
  ],
  "relations": []
}
\end{lstlisting}
\end{tcolorbox}

\subsection{Priority-driven Greedy Decomposition Protocol}

We employ a specialized decomposition algorithm to resolve semantic ambiguity in functional group identification. The protocol consists of three specific steps:

\paragraph{1. Priority Hierarchy Definition.}
We constructed a hierarchical dictionary where functional groups are ranked by heavy atom count, topological complexity, and semantic weight. High-complexity groups (\eg, Carboxyl $-\text{COOH}$, Amide $-\text{CONH}_2$) are assigned higher matching priority than their constituents (\eg, Carbonyl $C=O$, Hydroxyl $-OH$).

\paragraph{2. Recursive Matching with Exclusivity.}
For each SMILES string, we perform recursive substructure matching using RDKit \cite{landrum2013rdkit}. Crucially, we enforce an \textit{Atom-wise Exclusivity Constraint}: once an atom is assigned to a high-priority "super-node" (\eg, the Carbon in $-\text{COOH}$), it is locked and explicitly excluded from subsequent scans. This prevents the redundant labeling of a carboxyl carbon as a ketone, ensuring a bijective mapping between visual anchors and semantic tokens.

\paragraph{3. Residual Atom Handling.}
After the greedy matching process, any remaining atoms (typically satisfying the saturated alkane skeleton) are retained as atomic tokens. This \textit{hybrid-granularity} approach ensures that the model captures both high-level functional semantics and low-level structural details. We then calculate the 2D bounding box and the precise {anchor coordinates} $(x_{anc}, y_{anc})$ for each identified group.

\begin{tcolorbox}[colback=white, colframe=black, boxrule=0.5pt, arc=0mm, 
    title=\textbf{priority\_list\_file:}, fonttitle=\bfseries\small, left=2pt, right=2pt, top=2pt, bottom=2pt]
\scriptsize 
\begin{multicols}{2} 
\raggedright 
\begin{enumerate}[label=\arabic*., nosep, leftmargin=*, labelsep=3pt]
    \item Quaternary Ammonium
    \item Carboxylate
    \item Boronic Acid
    \item Boronate Ester
    \item Carboxylic Anhydride
    \item Orthoester
    \item Carbonate Ester
    \item Lactone
    \item Acetal/Ketal
    \item Hemiacetal/Hemiketal
    \item Epoxide
    \item Imide
    \item Lactam
    \item Carbamate
    \item Amidine
    \item Carboxylic Acid
    \item Ester
    \item Amide
    \item Acyl Halide
    \item Carbothioic O-Acid
    \item Carbothioic S-Acid
    \item Carbodithioic Acid
    \item Thiolester
    \item Thionoester
    \item Isocyanate
    \item Isothiocyanate
    \item Cyanate
    \item Thiocyanate
    \item Nitrile / Cyano
    \item Isonitrile
    \item Nitro
    \item Nitroso
    \item Oxime
    \item Azide
    \item Azo
    \item Hydrazine
    \item Primary Amine
    \item Secondary Amine
    \item Tertiary Amine
    \item Imine
    \item Sulfo (Sulfonic Acid)
    \item Sulfonate
    \item Sulfonyl
    \item Sulfino
    \item Disulfide
    \item Thiol/Sulfhydryl
    \item Sulfide(Thioether)
    \item Sulfinyl(Sulfoxide)
    \item Aldehyde
    \item Acetyl
    \item Ketone/Carbonyl
    \item Phenol
    \item Alcohol/Hydroxyl
    \item Hydroperoxy
    \item Peroxy
    \item Ether
    \item Methylenedioxy
    \item Aryl Halide
    \item Alkyl Halide
    \item Alkynyl
    \item Alkenyl
    \item Pyridyl
    \item Imidazole
    \item Aryl
    \item Carbonyl
    \item Halo
\end{enumerate}
\end{multicols}
\end{tcolorbox}


\section{OCRD-Bench Framework: Design and Metrics}
\label{sec:appendix_benchmark_metrics}

\begin{figure*}[t]
    \centering
    \includegraphics[width=0.95\textwidth]{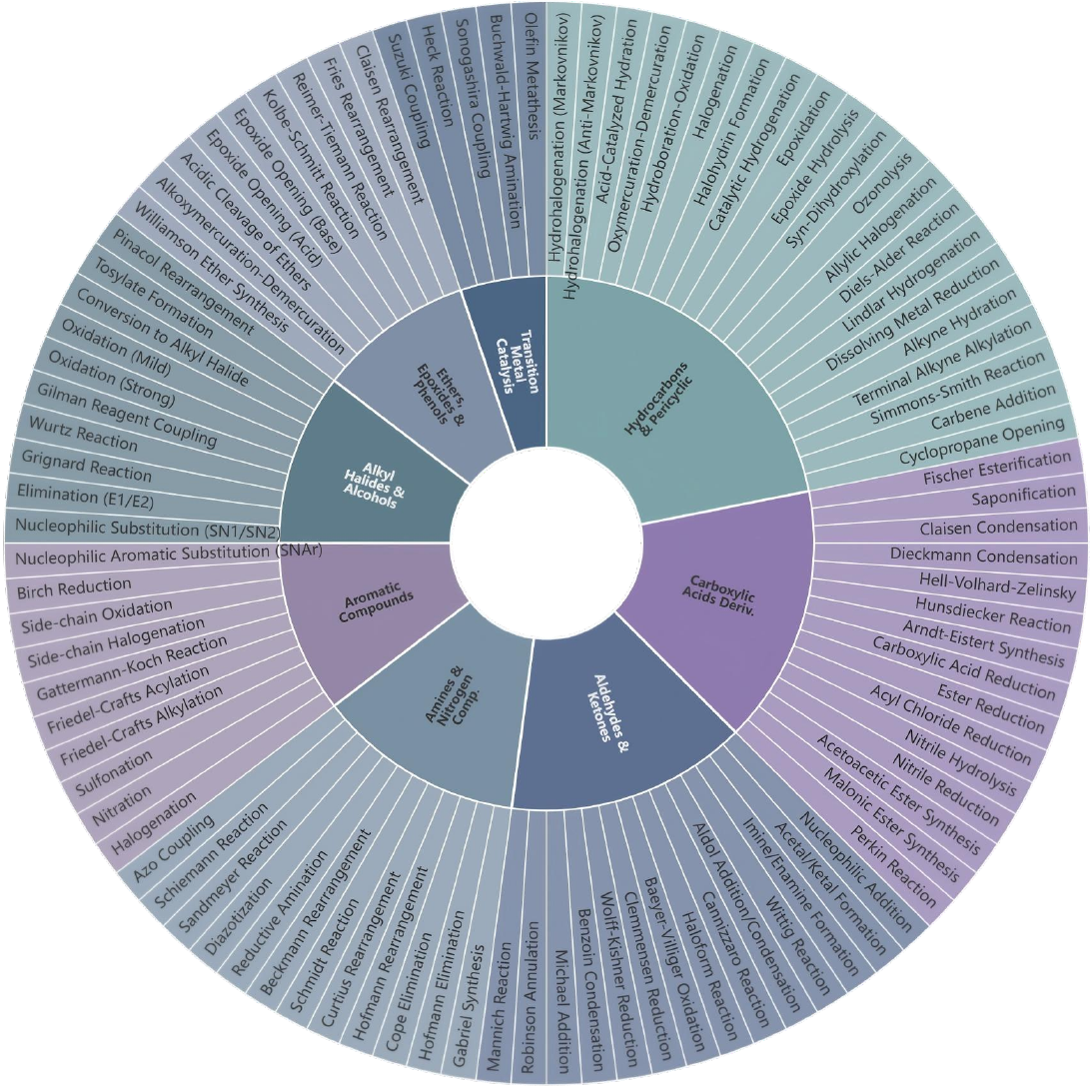}
    \caption{\textbf{Taxonomy of OCRD-Bench.} The benchmark spans 8 major categories of organic reactions, ranging from fundamental Hydrocarbons to complex Transition Metal Catalyzed Reactions. This diversity ensures a robust evaluation of model generalization across different chemical domains.}
    \label{fig:taxonomy}
\end{figure*}

To comprehensively evaluate Multimodal Large Language Models (MLLMs) on organic chemistry reasoning, we designed \textbf{OCRD-Bench}, a hierarchical benchmark covering 8 major reaction categories (see Figure~\ref{fig:taxonomy}). The evaluation is structured into three cognitive tiers, ranging from visual perception to deep mechanistic reasoning. Here, we detail the task design and the corresponding metric protocol for each level.

\subsection{Level 1: Visual Perception (L1)}
\textbf{Task Definition (Q1--Q2).} 
This tier tests the fundamental ability to ground visual features to symbolic representations. Taking \textit{Friedel-Crafts Alkylation} as an example, the model is prompted: \textit{"Identify the substrate (Benzene) and reagent (Chloroethane) in the image and encode them into SMILES."}

\textbf{Metric: Strict Structural Equivalence.} 
Standard text-generation metrics (e.g., Levenshtein distance) fail to capture chemical validity. Instead, we utilize RDKit to generate Morgan Fingerprints (radius=2, 2048 bits) for both the predicted ($P$) and ground-truth ($G$) SMILES. The score is calculated as a strict binary metric:
\begin{equation}
    \text{S}_{L1} = \mathbb{1}(\text{Tanimoto}(P, G) = 1.0) \times 100
\end{equation}
where $\mathbb{1}$ is the indicator function. This strict criterion ensures that any hallucinated atom, incorrect bond type, or stereochemical error results in a zero score, prioritizing topological fidelity over partial character matches.

\subsection{Level 2: Chemical Knowledge (L2)}
\textbf{Task Definition (Q3--Q4).} 
This tier assesses the retrieval of static knowledge regarding reactivity, selectivity (regio-/chemo-), and reaction conditions. Example task: \textit{"Compare the reactivity of chloroethane vs. chlorobenzene in this reaction system. Explain the scope and limitations regarding carbocation rearrangement."}

\textbf{Metric: LLM-as-a-Judge (Knowledge with Rationale).} 
While L2 tasks follow a single-choice or multiple-choice format, we do not rely solely on option accuracy to prevent "lucky guesses." We employ a strong proprietary model (Gemini 2.5 Pro) as an impartial judge to assess two components:
\begin{itemize}
    \item \textbf{Option Correctness:} Whether the selected answer matches the key.
    \item \textbf{Rationale Validity:} Whether the generated explanation logically derives the answer using correct chemical principles. 
\end{itemize}
A high score is awarded only when both the choice and the reasoning are correct.

\subsection{Level 3: Mechanistic Reasoning (L3)}
\textbf{Task Definition (Q5).} 
This tier tests deep reasoning capabilities. The model must generate a valid causal chain of intermediates rather than simply hallucinating the final product. Example task: \textit{"Provide the complete reaction mechanism, describing the electron flow and intermediate structures using SMILES sequences."}

\textbf{Metric: LLM-as-a-Judge (Completeness \& Consistency).} 
For open-ended mechanism generation, the judge compares the model's response against the standard exam key based on:
\begin{itemize}
    \item \textbf{Completeness:} Coverage of all key intermediates (e.g., formation of the Wheland intermediate) and electron flow steps.
    \item \textbf{Consistency:} Logical flow and absence of chemically impossible steps.
\end{itemize}
The judge assigns a scalar consistency score (0-100) reflecting the semantic alignment with the ground truth mechanism.

\subsection{Detailed Performance Data}
Table \ref{tab:ablation_breakdown} provides the fine-grained metrics for the baseline setting, serving as the reference point for the gains illustrated in the main text.

\begin{table}[h]
\centering
\caption{Detailed Breakdown of Baseline Performance (SMILES Only). This table lists the fine-grained metrics (Rec, Know, Reas) for the visual-only setting. The specific gains achieving the final Total Scores are visualized in Figure \ref{fig:ablation_chart}.}
\label{tab:ablation_breakdown}
\setlength{\tabcolsep}{3.5mm}
\resizebox{\columnwidth}{!}{%
\begin{tabular}{l|cccc}
\toprule
\multirow{2}{*}{\textbf{Model}} & \multicolumn{4}{c}{\textbf{SMILES Only}} \\
\cmidrule(lr){2-5} & Rec & Know & Reas & \textbf{Total} \\
\midrule
\multicolumn{5}{l}{\textit{\textbf{Proprietary SOTA Models}}} \\
\midrule
Gemini 2.5 Pro & 92.00 & 72.00 & 67.20 & 74.08 \\
GPT-5 (Preview) & 92.00 & 67.50 & 61.70 & 70.08 \\
Qwen3-Max & 92.00 & 61.00 & 59.40 & 66.56 \\
GPT-4o & 92.00 & 59.50 & 55.20 & 64.28 \\
Claude 4 Sonnet & 92.00 & 61.00 & 50.20 & 62.88 \\
\midrule
\multicolumn{5}{l}{\textit{\textbf{Open-Weights}}} \\
\midrule
Yi-Lightning & 92.00 & 44.00 & 40.20 & 52.08 \\
Llama-4-109B & 92.00 & 40.00 & 36.20 & 48.88 \\
Intern-S1 & 92.00 & 41.50 & 38.00 & 50.20 \\
ChemVLM & 92.00 & 16.50 & 15.00 & 31.00 \\
\bottomrule
\end{tabular}%
}
\end{table}

\begin{table}[h]
\centering
\caption{Detailed Performance Benchmark with Rxnscribe. This table lists the fine-grained metrics (Rec, Know, Reas) when using the external SOTA tool (RxnScribe) as the visual frontend. Note that the Recognition score is standardized to the tool's performance ceiling (86.0\%).}
\label{tab:molscribe_data}
\setlength{\tabcolsep}{3.5mm}
\resizebox{\columnwidth}{!}{%
\begin{tabular}{l|cccc}
\toprule
\multirow{2}{*}{\textbf{Model}} & \multicolumn{4}{c}{\textbf{RxnScribe}} \\
\cmidrule(lr){2-5} & Rec & Know & Reas & \textbf{Total} \\
\midrule
\multicolumn{5}{l}{\textit{\textbf{Proprietary SOTA Models}}} \\
\midrule
Gemini 2.5 Pro & 86.00 & 68.00 & 66.70 & 71.08 \\
GPT-5 (Preview) & 86.00 & 64.50 & 62.40 & 67.96 \\
Qwen3-Max & 86.00 & 66.00 & 61.20 & 68.08 \\
GPT-4o & 86.00 & 63.00 & 58.50 & 65.80 \\
Claude 4 Sonnet & 86.00 & 61.00 & 58.70 & 65.08 \\
\midrule
\multicolumn{5}{l}{\textit{\textbf{Open-Weights}}} \\
\midrule
Yi-Lightning & 86.00 & 54.50 & 51.20 & 59.48 \\
Llama-4-109B & 86.00 & 52.00 & 48.60 & 57.44 \\
Intern-S1 & 86.00 & 52.50 & 50.10 & 58.24 \\
ChemVLM & 86.00 & 29.50 & 28.50 & 40.40 \\
\bottomrule
\end{tabular}%
}
\end{table}


\begin{figure*}[t]
    \centering
    \includegraphics[width=\linewidth, height=20cm, keepaspectratio]{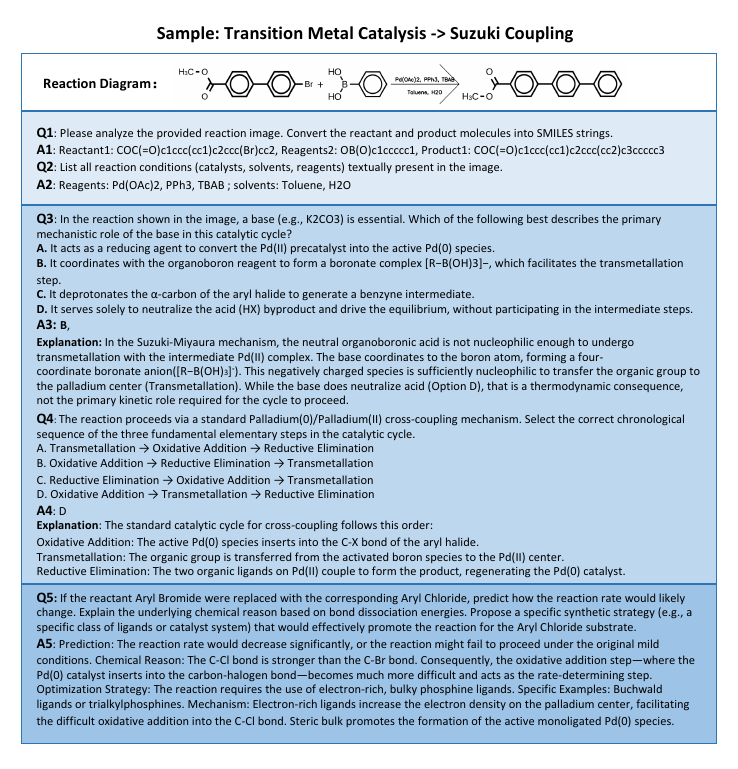}
    \caption{\textbf{An Example Case from OCRD-Bench.} We use the \textit{Suzuki-coupling} to illustrate our hierarchical evaluation design. \textbf{L1 (Q1, Q2)} assesses visual perception and structure grounding. \textbf{L2 (Q3, Q4)} evaluates static knowledge retrieval. \textbf{L3 (Q5)} tests deep mechanistic reasoning.}
    \label{fig:example}
    \vspace{3em}
\end{figure*}

\section{Case Study}
\label{app:detailed_data}

To demonstrate the efficacy of our method in handling complex chemical reasoning, we conduct a detailed qualitative analysis on a representative sample from the OCRD-Bench (Transition Metal Catalysis). We select the Suzuki-Miyaura Coupling reaction, a cornerstone of organic synthesis, which requires the model to navigate through a multi-step reasoning chain: from structural recognition to condition extraction, and finally to mechanistic inference.

Through this case study, we aim to highlight two critical failure modes prevalent in traditional Multimodal Large Language Models (MLLMs):

\begin{itemize}
    \item \textbf{Visual Deficit:} The inability to accurately resolve fine-grained structural details (e.g., bond orders, stereochemistry, and connectivity) in dense molecular diagrams.
    \item \textbf{Semantic Disconnect:} The representation-knowledge mismatch. Since LLMs acquire the vast majority of their chemical knowledge from textbooks—where entities are primarily referenced by names or aliases (e.g., "Aryl halide") rather than linear strings—SMILES often fails to effectively "awaken" (activate) the model's deep-seated domain knowledge.
\end{itemize}

\subsection{Visual Deficit in General-Purpose VLMs}
\label{ssec:case_study}
To explicitly demonstrate the \textit{Visual Deficit} discussed in Section~\ref{sec1}, we present a qualitative comparison using a Suzuki-Miyaura coupling reaction sample from the OCRD-Bench. As shown in Figure~\ref{fig:example}, the input image contains both dense molecular graphs (reactants/products) and textual reaction conditions.

\begin{tcolorbox}[
    colback=blue!3, 
    colframe=black!70, 
    arc=2mm, 
    left=3pt, 
    right=3pt,
    boxsep=5pt, 
    boxrule=0.5pt, 
    fonttitle=\bfseries, 
    title=Vanilla LLM: Qwen3-Max,
    breakable
]

{\small \textit{\textbf{Context:} Single-step Reaction Parsing Prompt}}
\vspace{0.2cm}
\hrule
\vspace{0.2cm}
\textbf{Q1, Q2}\\
\textbf{System Prompt:}

You are an expert organic chemist and an advanced Optical Chemical Structure Recognition (OCSR) system. Your task is to accurately extract chemical information from reaction scheme images.

\vspace{0.3cm}

\textbf{User Prompt:}

Please analyze the attached image of a chemical reaction. Your goal is to identify the chemical structures and the text conditions, and extract them into a structured format.

\vspace{0.1cm}
\textbf{Instructions:}
\begin{itemize}
    \item \textbf{Reactants:} Identify the molecules to the left of the reaction arrow. Convert their structures into SMILES. If there are multiple reactants, separate them with a period (.).
    \item \textbf{Products:} Identify the molecules to the right of the reaction arrow. Convert their structures into SMILES.
    \item \textbf{Conditions:} Read the text above and below the reaction arrow. Based on your chemical knowledge, categorize these substances into "Reagents" (including catalysts, bases, ligands, additives) and "Solvents".
\end{itemize}

\vspace{0.1cm}
\textbf{Output Format:}
Please strictly follow this format (do not output markdown code blocks, just the raw text):

\begin{quote}
\ttfamily \raggedright

\hangindent=6em \hangafter=1
Reactants: [SMILES string]

\hangindent=6em \hangafter=1
Products: [SMILES string]

\hangindent=6em \hangafter=1
Reagents: [List of reagents, separated by comma]

\hangindent=6em \hangafter=1
Solvents: [List of solvents, separated by comma]
\end{quote}

\vspace{0.2cm}
\hrule
\vspace{0.2cm}

\textbf{Answer:}

\texttt{Reactants: \seqsplit{COC(=O)c1ccc(c2cc(Br)ccc2c1)c1ccccc1.OB(O)(O)c1ccccc1} \\
Products: \seqsplit{COC(=O)c1ccc(c2ccccc2c1)c1ccccc1} \\
Reagents: Pd(OAc)2, PPh3, TBAB \\
Solvents: Toluene, H2O}

\end{tcolorbox}

\begin{tcolorbox}[
    colback=blue!3, 
    colframe=black!70, 
    arc=2mm, 
    left=3pt, 
    right=3pt,
    boxsep=5pt, 
    boxrule=0.5pt, 
    fonttitle=\bfseries, 
    title=FG-VLM Answer,
    breakable
]
\textbf{Answer:}

\texttt{Reactants: \seqsplit{COC(=O)c1ccc(-c2ccc(Br)cc2)cc1.OB(O)c1ccccc1} \\
Products: \seqsplit{COC(=O)c1ccc(-c2ccc(-c3ccccc3)cc2)cc1} \\
Reagents: Pd(OAc)2, PPh3, TBAB \\
Solvents: Toluene, H2O}

\end{tcolorbox}

\paragraph{The Illusion of Understanding.}
General-purpose VLMs, such as Qwen3-max, exhibit a deceptive capability profile. As observed in the detailed model outputs, Qwen3-max successfully retrieves the textual reaction conditions, correctly identifying the catalyst system (``Pd(OAc)$_2$, PPh$_3$'') and solvents (``Toluene, H$_2$O''). This aligns with the model's strong OCR capabilities for standard text. 

However, a critical failure occurs in the structural interpretation of the molecular components (Q1). The ground truth reactant is a biaryl bromide (\textit{methyl 4'-(bromophenyl)-benzoate}), composed of two linearly connected benzene rings. In contrast, Qwen3-max generates a hallucinatory SMILES string: 
\begin{center}
\texttt{COC(=O)c1ccc(c2cc(Br)ccc2c1)c1ccccc1}
\end{center}
This output reveals a severe topological error: it incorrectly introduces an additional phenyl ring (``\texttt{c1ccccc1}'') and disrupts the biaryl connectivity, effectively fabricating a molecule that does not exist in the image. This exemplifies the \textit{Visual Deficit}: the model perceives the visual tokens but fails to resolve the strict graph topology, resulting in structural hallucination.

\paragraph{Resolving Connectivity via Visual Anchors.}
In contrast, our FG-vLM correctly reconstructs the reactant SMILES by leveraging the proposed \textit{Hybrid-Granularity Visual Anchor Modeling}. Instead of treating the molecule as a sequence of atomic characters, FG-vLM identifies the rigid functional groups (e.g., the phenyl rings and ester group) as stable visual anchors. By explicitly modeling the directional vectors between these anchors, our method enforces topological fidelity, successfully recognizing the correct two-ring system and the precise position of the bromine substituent. This case underscores that bridging the visual deficit requires more than scaling generic vision encoders; it demands domain-specific grounding of chemical substructures.

\subsection{Beyond Correct Answers: Grounding Reasoning in Visual Reality}
\label{ssec:reasoning_analysis}

While quantitative metrics (Accuracy) indicate the final performance, a qualitative inspection of the reasoning process reveals the profound impact of our \OURS{} framework in bridging the \textit{Semantic Disconnect}. We analyze two representative knowledge tasks (Q3 and Q4) to demonstrate the difference between \textit{parametric retrieval} and \textit{visually grounded reasoning}.

\begin{tcolorbox}[
    colback=blue!3, 
    colframe=black!70, 
    arc=2mm, 
    left=3pt, 
    right=3pt,
    boxsep=5pt, 
    boxrule=0.5pt, 
    fonttitle=\bfseries, 
    title=Know Task,
    breakable
]

{\small \textit{\textbf{Context:} Prompt with Knowledge Injection}}
\vspace{0.2cm}
\hrule
\vspace{0.2cm}

\textbf{System Prompt:}

You are an expert organic chemist and academic researcher. Your task is to analyze chemical reaction images and answer reasoning questions based on your domain knowledge of reaction mechanisms, reagents, and catalytic cycles.

\textbf{CRITICAL OUTPUT RULES:}
\begin{enumerate}
    \item You must output the response STRICTLY in the following format.
    \item Do not include any introductory text or concluding remarks.
    \item For "answer": Output the single letter (e.g., A) or letters separated by commas.
    \item For "explanation": Provide a concise but professional chemical justification.
\end{enumerate}

\textbf{Standard Output Format:}
\texttt{answer: [Option Letter(s)]}\\
\texttt{explanation: [Your detailed reasoning content]}

\vspace{0.2cm}
\hrule
\vspace{0.2cm}

\textbf{User Prompt:}

\textbf{Task:} Analyze the provided chemical reaction image and the supplementary information, and then answer the following question.

\textbf{Supplementary information}:[]

\textbf{Question:}[]

\textbf{Question Type:}[]

\vspace{0.2cm}
\hrule
\vspace{0.2cm}


\textbf{\large SMILES Only}

\textbf{Supplementary Information (Input):}
{\footnotesize \texttt{\seqsplit{Reactant1: COC(=O)c1ccc(-c2ccc(Br)cc2)cc1, Reagents2: OB(O)c1ccccc1, Product1: COC(=O)c1ccc(-c2ccc(-c3ccccc3)cc2)cc1, Reagents: Pd(OAc)2, PPh3, TBAB ; solvents: Toluene, H2O}}}

\textbf{Q3}

\textbf{Model Output:}
\begin{quote}
\ttfamily \raggedright
answer: B \\
explanation: The reaction depicted is a Suzuki-Miyaura cross-coupling between an aryl bromide and phenylboronic acid, catalyzed by Pd(OAc)$_2$ with PPh$_3$ as ligand and TBAB as phase-transfer catalyst in a toluene/water mixture. Although the base (e.g., K$_2$CO$_3$) is not explicitly shown in the diagram, it is essential for the catalytic cycle. The primary role of the base is to deprotonate the boronic acid, forming a more nucleophilic boronate species [R-B(OH)$_3$]$^-$, which facilitates the transmetallation step—where the aryl group from the palladium complex transfers to the boronate, enabling coupling. This step is crucial for efficient coupling. Option A is incorrect because the reduction of Pd(II) to Pd(0) is typically achieved by the ligands or solvent, not the base. Option C is incorrect because benzyne formation is not involved; this is not a elimination-based mechanism. Option D is incorrect because the base actively participates in the transmetallation step, not merely neutralizing acid. Thus, B is correct.
\end{quote}

\vspace{0.2cm}
\hrule
\vspace{0.2cm}

\textbf{Q4}

\textbf{Model Output:}
\begin{quote}
\ttfamily \raggedright
answer: D \\
explanation: The reaction shown is a Suzuki-Miyaura cross-coupling, which proceeds via a Pd(0)/Pd(II) catalytic cycle. The correct chronological sequence of elementary steps is: (1) Oxidative Addition—where the Pd(0) species adds across the C--Br bond of the aryl bromide to form a Pd(II) complex; (2) Transmetallation—where the organoboron reagent (phenylboronic acid, activated by base) transfers its phenyl group to the palladium center, displacing the halide; and (3) Reductive Elimination—where the two aryl groups on palladium couple to form the biaryl product and regenerate the Pd(0) catalyst. This sequence is well-established in palladium-catalyzed cross-coupling reactions. Option A is incorrect because transmetallation cannot occur before oxidative addition, as no Pd--C bond exists yet. Option B reverses the order of transmetallation and reductive elimination, which is mechanistically invalid. Option C starts with reductive elimination, which requires prior formation of the Pd(II) intermediate, making it impossible as the first step. Thus, D is the only correct sequence.

\end{quote}

\vspace{0.2cm}
\hrule
\vspace{0.2cm}

\textbf{\large ChemVA}

\textbf{Supplementary Information (Input):}
{\footnotesize \texttt{\seqsplit{Reactant1: COC(=O)c1ccc(-c2ccc(Br)cc2)cc1, Synonyms: methyl 4'-bromo[1,1'-biphenyl]-4-carboxylate, IUPAC Name: methyl 4-(4-bromophenyl)benzoate, Reagents2: OB(O)c1ccccc1, Synonyms: Phenylboronic acid, IUPAC Name: phenylboronic acid, Product1: COC(=O)c1ccc(-c2ccc(-c3ccccc3)cc2)cc1, Synonyms: , IUPAC Name: methyl 4-(4-phenylphenyl)benzoate, Reagents: Pd(OAc)2, PPh3, TBAB ; solvents: Toluene, H2O}}}

\textbf{Q3}

\textbf{Model Output:}
\begin{quote}
\ttfamily \raggedright
answer: B \\
explanation: The reaction shown is a Suzuki–Miyaura cross-coupling between methyl 4-(4-bromophenyl)benzoate and phenylboronic acid, catalyzed by Pd(OAc)$_2$/PPh$_3$ in the presence of TBAB, under biphasic conditions (toluene/H$_2$O). In this catalytic cycle, the base (e.g., K$_2$CO$_3$) plays a critical role during the transmetallation step. It deprotonates the boronic acid to form a more nucleophilic boronate species [R-B(OH)$_3$]$^-$, which facilitates the transfer of the organic group from boron to palladium. This step is essential because the neutral boronic acid is less reactive toward transmetallation with the Pd(II) or Pd(0) intermediate. Option A is incorrect because the reduction of Pd(II) to Pd(0) is typically facilitated by phosphine ligands or solvent-mediated processes, not the base. Option C is incorrect as benzyne formation requires strong bases and specific substrates (e.g., ortho-dihalides), which are not present here. Option D is incorrect because the base is not merely a stoichiometric neutralizer; it actively participates in the mechanism via coordination and activation of the boronic acid. Thus, the primary mechanistic role of the base is to generate the reactive boronate complex, making option B correct.
\end{quote}

\vspace{0.2cm}
\hrule
\vspace{0.2cm}

\textbf{Q4}

\textbf{Model Output:}
\begin{quote}
\ttfamily \raggedright
answer: D \\
explanation: The reaction depicted is a Suzuki-Miyaura cross-coupling between an aryl bromide (methyl 4-(4-bromophenyl)benzoate) and phenylboronic acid, catalyzed by Pd(OAc)$_2$ in the presence of PPh$_3$ and TBAB in toluene/water. This transformation proceeds via a well-established Pd(0)/Pd(II) catalytic cycle. The correct chronological sequence of elementary steps is: (1) Oxidative Addition, where the Pd(0) species adds across the C--Br bond of the aryl halide to form a Pd(II) intermediate; (2) Transmetallation, in which the organoboron reagent transfers its phenyl group to the palladium center, displacing the bromide and forming a Pd(II)-aryl complex; and (3) Reductive Elimination, where the two aryl groups coupled on the Pd(II) center eliminate to form the biaryl product and regenerate Pd(0). Option D correctly lists this sequence: Oxidative Addition $\to$ Transmetallation $\to$ Reductive Elimination. Options A, B, and C are incorrect because they misorder these essential mechanistic steps—transmetallation cannot occur before oxidative addition, and reductive elimination must follow transmetallation to form the new C--C bond.
\end{quote}

\end{tcolorbox}

\paragraph{Q3: Mechanistic Fidelity vs. Hallucinated Logic.}
In Question 3 (Role of the Base), both the baseline (SMILES Only) and \OURS{} correctly selected Option B. However, a close examination of the generated explanations reveals a critical divergence in chemical understanding.
\begin{itemize}
    \item The Baseline's Failure: Although the baseline selected the correct option, its explanation contains a fundamental mechanistic error. It states: \textit{``the aryl group from the palladium complex transfers to the boronate.''} This is chemically inverted; in Transmetallation, the organic group transfers \textit{from} the activated boron species \textit{to} the palladium center. This ``Right Answer, Wrong Reason'' phenomenon suggests the baseline relies on superficial keyword matching or hallucinated logic rather than a true understanding of the catalytic cycle.
    \item ChemVA's Precision: In contrast, \OURS{} accurately describes the directionality of the group transfer (Boron $\to$ Palladium) and correctly identifies the nucleophilic activation process. This confirms that our Visual Anchor mechanism not only aids in structure recognition but also enforces logical consistency in downstream reasoning.
\end{itemize}

\paragraph{Q4: Instance-Level Awareness vs. Generic Templates.}
Question 4 asks for the sequence of the catalytic cycle. While the sequence is standard for all Suzuki couplings, the quality of the response depends on how well the model integrates the specific context of the input image.
\begin{itemize}
    \item Generic Retrieval (Baseline): The baseline refers to the reactant simply as an ``aryl bromide.'' While factually true, this generic label implies the model is retrieving a pre-trained template for ``Suzuki Reaction'' without attending to the specific molecular details shown in the image.
    \item Contextual Grounding (\OURS{}): \OURS{} exhibits superior context awareness by explicitly naming the complex substrate: \textit{``methyl 4-(4-bromophenyl)benzoate.''} This precision stems directly from our Semantic Activation mechanism, which translates visual features into explicit textual anchors (\textit{e.g., IUPAC names and synonyms}) before reasoning. It also correctly enumerates the specific reaction conditions (Pd(OAc)$_2$, PPh$_3$, TBAB) visible in the image. This proves that \OURS{} is not merely reciting a textbook definition but is performing \textit{instance-specific reasoning} grounded in the visual evidence extracted via the Semantic Activation module.
\end{itemize}

\begin{tcolorbox}[
    colback=blue!3, 
    colframe=black!70, 
    arc=2mm, 
    left=3pt, 
    right=3pt,
    boxsep=5pt, 
    boxrule=0.5pt, 
    fonttitle=\bfseries, 
    title=Reas Task,
    breakable
]

{\small \textit{\textbf{Context:} Open-Ended Reasoning Prompt}}
\vspace{0.2cm}
\hrule
\vspace{0.2cm}

\textbf{System Prompt:}

You are an expert organic chemist and synthetic consultant. Your task is to analyze chemical reaction images and answer advanced reasoning questions regarding substrate scope, reaction kinetics, and condition optimization.

\textbf{CRITICAL OUTPUT RULES:}
\begin{enumerate}
    \item You must output the response STRICTLY in the following format.
    \item Do not include any introductory text.
    \item For "answer": Provide a concise, direct conclusion (1-2 sentences).
    \item For "explanation": Provide a comprehensive analysis including Theoretical Basis, Mechanism Analysis, and Synthetic Strategy.
\end{enumerate}

\textbf{Standard Output Format:}
\texttt{answer: [Direct Conclusion/Prediction]}\\
\texttt{explanation: [Detailed Mechanism Analysis and Synthetic Proposal]}

\vspace{0.2cm}
\hrule
\vspace{0.2cm}

\textbf{User Prompt:}

\textbf{Task:} Analyze the provided chemical reaction image. Based on the specific reaction type and conditions shown, answer the following hypothetical reasoning question.

\textbf{Question:}[]

\textbf{Supplementary information:}[]

\textbf{Question Type:} Open-Ended Reasoning / Reaction Optimization

\vspace{0.2cm}
\hrule
\vspace{0.2cm}

\textbf{\large SMILES Only}

\textbf{Supplementary Information (Input):}
{\footnotesize \texttt{\seqsplit{Reactant1: COC(=O)c1ccc(-c2ccc(Br)cc2)cc1, Reagents2: OB(O)c1ccccc1, Product1: COC(=O)c1ccc(-c2ccc(-c3ccccc3)cc2)cc1, Reagents: Pd(OAc)2, PPh3, TBAB ; solvents: Toluene, H2O}}}

\textbf{Model Output:}
\begin{quote}
\ttfamily \raggedright
answer: The reaction rate would decrease significantly, or the reaction might fail to proceed under the original mild conditions. Chemical Reason: The C-Cl bond is stronger than the C-Br bond. Consequently, the oxidative addition step—where the Pd(0) catalyst inserts into the carbon-halogen bond—becomes much more difficult and acts as the rate-determining step.

explanation: \\
Theoretical Basis: The bond dissociation energy (BDE) of a C--Cl bond ($\sim$339 kJ/mol) is significantly higher than that of a C--Br bond ($\sim$276 kJ/mol), making the cleavage of the C--Cl bond in oxidative addition less favorable thermodynamically and kinetically. This increased bond strength results in a higher activation energy barrier for the oxidative addition step, which is the rate-determining step in Suzuki-Miyaura coupling reactions.

Mechanism Analysis: In the Suzuki coupling, the Pd(0) species undergoes oxidative addition with the aryl halide. For bromides, this step proceeds readily due to the weaker C--Br bond and better polarizability. However, when the bromine is replaced by chlorine, the stronger C--Cl bond resists oxidative addition, leading to slow or incomplete reaction. Additionally, the lower nucleophilicity and poorer leaving-group ability of chloride compared to bromide further hinder the process.

Synthetic Strategy: To overcome this challenge, electron-rich and sterically bulky phosphine ligands are essential. These ligands increase the electron density at the palladium center, enhancing its nucleophilicity and facilitating oxidative addition into the strong C--Cl bond. Specific examples include Buchwald-type ligands such as SPhos, XPhos, or RuPhos, which are known for their high activity in aryl chlorides. Alternatively, N-heterocyclic carbene (NHC) ligands like IPr or SIPr can be used, as they provide strong $\sigma$-donation and stability to the Pd(0) complex. The use of a more active catalyst system such as Pd(OAc)$_2$ with these advanced ligands, along with elevated temperatures or longer reaction times, may be required to achieve satisfactory yields.
\end{quote}

\vspace{0.2cm}
\hrule
\vspace{0.2cm}

\textbf{\large ChemVA (Ours)}

\textbf{Supplementary Information (Input):}
{\footnotesize \texttt{\seqsplit{Reactant1: COC(=O)c1ccc(-c2ccc(Br)cc2)cc1, Synonyms: methyl 4'-bromo[1,1'-biphenyl]-4-carboxylate, IUPAC Name: methyl 4-(4-bromophenyl)benzoate, Reagents2: OB(O)c1ccccc1, Synonyms: Phenylboronic acid, IUPAC Name: phenylboronic acid, Product1: COC(=O)c1ccc(-c2ccc(-c3ccccc3)cc2)cc1, Synonyms: , IUPAC Name: methyl 4-(4-phenylphenyl)benzoate, Reagents: Pd(OAc)2, PPh3, TBAB ; solvents: Toluene, H2O}}}

\textbf{Model Output:}
\begin{quote}
\ttfamily \raggedright
answer: The reaction rate would decrease significantly, or the reaction might fail to proceed under the original mild conditions. Chemical Reason: The C-Cl bond is stronger than the C-Br bond. Consequently, the oxidative addition step—where the Pd(0) catalyst inserts into the carbon-halogen bond—becomes much more difficult and acts as the rate-determining step.

explanation: \\
Theoretical Basis: The bond dissociation energy (BDE) of a C--Cl bond ($\sim$339 kJ/mol) is significantly higher than that of a C--Br bond ($\sim$285 kJ/mol), making oxidative addition of Pd(0) into a C--Cl bond thermodynamically less favorable and kinetically slower. This increased bond strength directly impacts the rate of the first step in the Suzuki-Miyaura coupling mechanism, which is oxidative addition. Additionally, chlorine is a poorer leaving group than bromine due to its lower polarizability and greater electronegativity, further reducing the rate of this step.

Mechanism Analysis: In the standard Suzuki coupling mechanism, oxidative addition is typically the rate-determining step when using aryl halides. With an aryl chloride instead of an aryl bromide, the electron-deficient palladium center struggles to cleave the stronger C--Cl bond. The PPh$_3$ ligand used in the original conditions is moderately electron-rich but not sufficiently so to stabilize the transition state for oxidative addition of a C--Cl bond. Moreover, PPh$_3$ has a moderate cone angle ($\sim$145$^\circ$), which may not favor formation of the active monoligated Pd(0) species required for efficient oxidative addition.

Synthetic Strategy: To overcome this limitation, electron-rich and sterically bulky phosphine ligands must be employed. Buchwald-type ligands such as SPhos, XPhos, or RuPhos are ideal because they increase electron density at the palladium center, facilitating oxidative addition of the strong C--Cl bond. These ligands also possess large cone angles ($>$150$^\circ$), promoting the formation of low-coordinate, highly reactive Pd(0) species. Alternatively, N-heterocyclic carbene (NHC) ligands like IPr or SIPr can be used due to their strong $\sigma$-donating ability, enhancing the nucleophilicity of Pd(0). A recommended optimized system would include Pd(OAc)$_2$ with XPhos (or SPhos) in a mixture of dioxane/water or toluene/water at elevated temperature (80--100\,$^\circ$C) to achieve successful coupling.
\end{quote}

\end{tcolorbox}

Question 5 serves as a stress test for synthetic planning, asking the model to predict reactivity changes for an Aryl Chloride substrate and propose an optimization strategy.
\begin{itemize}
    \item Baseline: The baseline correctly identifies the primary thermodynamic barrier: the higher Bond Dissociation Energy (BDE) of C--Cl vs. C--Br. However, its reasoning is \textit{unidirectional}, focusing solely on the substrate's limitations. It fails to explain specifically why the current ligand (PPh$_3$) is insufficient beyond generic electronic arguments, offering only a standard textbook response.
    \item ChemVA: It analyzes not just the substrate, but the specific incompatibility between the substrate and the ligand's steric properties.
    \begin{enumerate}
        \item \textit{Mechanistic Depth:} It explicitly cites the Cone Angle ($\sim$145$^\circ$) of PPh$_3$, reasoning that it is insufficient to promote the formation of the highly active \textit{monoligated} Pd(0) species required for difficult oxidative additions. This incorporates advanced organometallic concepts (steric bulk vs. coordination number) absent in the baseline.
        \item \textit{Operational Precision:} While the baseline suggests generic ligand classes, \OURS{} proposes a complete experimental protocol, including specific solvents (dioxane/water) and temperature ranges (80--100\,$^\circ$C). This reflects a "tactical" understanding of reaction conditions that is crucial for wet-lab success.
    \end{enumerate}
\end{itemize}

\subsection{The Role of Semantic Activation}
\label{ssec:semantic_activation_role}
The qualitative superiority of \OURS{} across Q3 (Mechanism), Q4 (Context), and Q5 (Optimization) highlights the critical function of \textbf{Semantic Activation}. 

Visual encodings (or SMILES strings) alone often trap Large Language Models in a "surface-level retrieval" mode, where they generate generic facts based on pattern matching. By translating visual features into explicit, knowledge-rich textual descriptors (\textit{e.g., IUPAC names, specific reagents like PPh$_3$}), our method unlocks the model's latent "expert subspace." 

For instance, in Q5, the token "PPh$_3$"—when explicitly presented via Semantic Activation—triggers associated concepts like \textit{Cone Angles} and \textit{Tolman parameters} from the model's pre-training corpus. This allows the model to move beyond simple property retrieval (BDE values) to complex, multi-variable reasoning (steric-electronic fits), effectively bridging the \textit{Semantic Disconnect} between visual perception and deep chemical reasoning.

\end{document}